\documentclass[sigconf]{aamas}

\usepackage{algorithm}
\usepackage[noend]{algpseudocode}
\usepackage{amsthm}
\usepackage{mathrsfs}
\usepackage{bbm}
\usepackage{appendix}
\usepackage{multirow}
\usepackage{multicol} 
\usepackage{array}
\usepackage{booktabs}
\usepackage{makecell}
\usepackage{subcaption}
\usepackage{enumitem}
\usepackage{nicefrac}

\newtheorem{definition}{Definition}

\usepackage{xspace}
\usepackage{cleveref}
\usepackage{newfloat}
\usepackage{listings}
\DeclareCaptionStyle{ruled}{labelfont=normalfont,labelsep=colon,strut=off} 

\floatstyle{ruled}
\newfloat{listing}{tb}{lst}{}
\floatname{listing}{Listing}

\newcommand{\rfpg}{\textsc{rfPG}\xspace}
\newcommand{\toolname}{\textsc{Lexpop}\xspace}

\newcommand{\saynt}{\textsc{Saynt}\xspace}
\newcommand{\paynt}{\textsc{Paynt}\xspace}

\newcommand{\storm}{\textsc{Storm}\xspace}
\newcommand{\alergia}{\textsc{Alergia}\xspace}
\newcommand{\sig}{\textsc{SIG}\xspace}

\newcommand{\mdp}{M}
\newcommand{\pomdp}{\mathcal{M}}
\newcommand{\hmpomdp}{\mathscr{M}}
\newcommand{\trsfun}{T}
\newcommand{\rewfun}{R}
\newcommand{\obsset}{Z}
\newcommand{\obsfun}{O}
\newcommand{\actset}{A}
\newcommand{\obs}{z}

\newcommand{\fsc}{\pi_{F}}
\newcommand{\FSC}{\mathcal{F}}
\newcommand{\actfun}{\delta}
\newcommand{\memfun}{\eta}
\newcommand{\actmemfun}{\sigma}

\newcommand{\hiddenstate}{h}
\newcommand{\rnn}{\pi_{\text{RNN}_{\theta}}}

\newcommand{\distr}{\Delta}

\newtheorem{problem}{Problem}
\newcommand{\tuple}[1]{\langle #1 \rangle}
\newcommand{\softmax}{\text{Softmax}}

\newcommand{\pomdpbatch}{B}

\newcommand{\statesn}{{\lvert N\rvert}}

\DeclareMathOperator*{\argmax}{argmax}

\usepackage{wrapfig}
\usepackage{balance} 

\doi{TBFQ5922}

\makeatletter
\gdef\@copyrightpermission{
  \begin{minipage}{0.2\columnwidth}
   \href{https://creativecommons.org/licenses/by/4.0/}{\includegraphics[width=0.90\textwidth]{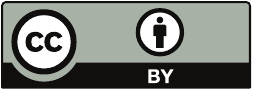}}
  \end{minipage}\hfill
  \begin{minipage}{0.8\columnwidth}
   \href{https://creativecommons.org/licenses/by/4.0/}{This work is licensed under a Creative Commons Attribution International 4.0 License.}
  \end{minipage}
  \vspace{5pt}
}
\makeatother

\setcopyright{ifaamas}
\acmConference[AAMAS '26]{Proc.\@ of the 25th International Conference
on Autonomous Agents and Multiagent Systems (AAMAS 2026)}{May 25 -- 29, 2026}
{Paphos, Cyprus}{C.~Amato, L.~Dennis, V.~Mascardi, J.~Thangarajah (eds.)}
\copyrightyear{2026}
\acmYear{2026}
\acmDOI{}
\acmPrice{}
\acmISBN{}

\acmSubmissionID{623}

\title{
Finite-State Controllers for 
(Hidden-Model) 
POMDPs\\ using Deep Reinforcement Learning}
\author{David Hud\'{a}k}
\affiliation{
  \orcid{0009-0003-9158-3407}
  \institution{Brno University of Technology}
  \city{Brno}
  \country{Czech Republic}}
\email{ihudak@fit.vutbr.cz}

\author{Maris F. L. Galesloot}
\affiliation{
  \orcid{0009-0002-5112-8584}
  \institution{Radboud University}
  \city{Nijmegen}
  \country{The Netherlands}}
\email{maris.galesloot@ru.nl}

\author{Martin Tappler}
\affiliation{
  \orcid{0000-0002-4193-5609}
  \institution{TU Wien}
  \city{Vienna}
  \country{Austria}} 
\email{martin.tappler@tuwien.ac.at}

\author{Martin Kurečka}
\affiliation{
  \orcid{0000-0001-5092-9190}
  \institution{Masaryk University}
  \city{Brno}
  \country{Czech Republic}}
\email{martin.kurecka@fi.muni.cz}

\author{Nils Jansen}
\affiliation{
  \orcid{0000-0003-1318-8973}
  \institution{Ruhr-Universtät Bochum \& Radboud~University}
  \city{Bochum}
  \country{Germany}
  }
\email{n.jansen@rub.de}

\author{Milan \v{C}e\v{s}ka}
\affiliation{
  \orcid{0000-0002-0300-9727}
  \institution{Brno University of Technology}
  \city{Brno}
  \country{Czech Republic}}
\email{ceskam@fit.vutbr.cz}

\begin{abstract}

Solving partially observable Markov decision processes (POMDPs) requires computing policies under imperfect state information.
Despite recent advances, the scalability of existing POMDP solvers remains limited.
Moreover, many settings require a policy that is \emph{robust} across multiple POMDPs, further aggravating the scalability issue. 
We propose the \toolname framework 
for POMDP solving.
\toolname (1) employs deep reinforcement learning to train a \emph{neural policy}, represented by a \emph{recurrent neural network}, and (2) constructs a \emph{finite-state controller} mimicking the neural policy through efficient extraction methods. 
Crucially, unlike neural policies, such controllers can be formally evaluated, providing performance guarantees. 
We extend \toolname to compute robust policies for \emph{hidden-model} POMDPs~(HM-POMDPs), which describe finite sets of POMDPs. 
We associate every extracted controller with its worst-case POMDP.
Using a set of such POMDPs, 
we 
iteratively train a robust neural policy and consequently extract a robust controller.
Our experiments show that on problems with large state spaces, \toolname outperforms state-of-the-art solvers for POMDPs as well as HM-POMDPs.

\end{abstract}


\newcommand{\BibTeX}{\rm B\kern-.05em{\sc i\kern-.025em b}\kern-.08em\TeX}

\begin{document}


\pagestyle{fancy}
\fancyhead{}


\maketitle 




\section{Introduction}\label{introduction}
Partially observable Markov decision processes (POMDPs) \cite{kaelbling1998planning} are a ubiquitous model for automated sequential decision-making under imperfect state information.
Applications for POMDPs range from robotics~\cite{annurev:/content/journals/10.1146/annurev-control-042920-092451,kurniawati2008sarsop}, over network protocols~\cite{5683311}, to healthcare~\cite{healthcare10020283}.
Solving a given discrete POMDP, i.e., computing policies that maximize expected rewards,  
is intractable in general~\cite{madani2000undecidability}, as optimal policies may require an infinite memory. 
State-of-the-art \emph{model-based} POMDP solvers thus leverage various approximation techniques or heuristics.
Here, the primary focus is on (1) the analysis of belief states representing probability distributions over the unobservable states~\cite{smith2004heuristic,kurniawati2008sarsop,uai24} and (2) symbolic search techniques over finite-state policies~\cite{kumar2015history,andriushchenko23search}.
Furthermore, since exact environment models are rarely available, recent research has considered computing \emph{robust} policies that provide sufficient performance over a given set of POMDPs~\cite{DBLP:conf/aaai/Cubuktepe0JMST21,hmpomdps,pip-rnn}. 
A joint limitation of the above model-based methods is their scalability with respect to the model size.
These methods must either (1) unfold a very large belief space or (2)~explore a very large space of potential finite-state policies.

In contrast to model-based methods, model-free deep reinforcement learning~(DRL) provides a means to mitigate these scalability challenges.
DRL for POMDPs uses \emph{recurrent neural networks}~(RNNs) to represent memory,
allowing policies to generalize behavior for similar states and observations using only interactions with the environment~\cite{SuttonB:rlbook,Kochenderfer2022}. 
While DRL scales well to large state spaces, there are two crucial obstacles.
First, aligning DRL with our goal of solving a given (set of) POMDPs presents a significant technical challenge, as DRL must be set up correctly to achieve well-performing policies, requiring a highly efficient pipeline, combined with insights from the literature on addressing issues of sparse rewards, robustness, and partial observability~\cite{challenges_rl,DBLP:conf/icml/NiES22}. 
To the best of our knowledge, such a pipeline using DRL for solving discrete POMDPs has not yet been achieved.
Second, the exact evaluation of DRL policies is not straightforward.
The resulting lack of guarantees regarding the performance of a policy restricts the utilization of DRL to games~\cite{DBLP:journals/nature/WurmanBKMS0CDE022} 
or problems where soft guarantees are sufficient, such as language models~\cite{rlhf_2022}.
Using POMDPs in safety-critical domains requires means to exactly evaluate the resulting policies. 

 \paragraph{Contributions.}
 Our main contribution is \toolname~-- Learning-based policy EXtraction for POMDP Planning -- 
 that yields high-quality finite-state policies with verified performance. 
\toolname builds on two key insights: (1)~DRL, when engineered correctly, scales up to environments that are beyond reach of existing POMDP solvers. 
 (2)~Finite-state approximations of RNN-based policies often achieve comparable performance while enabling computationally tractable model-based evaluation, providing the required guarantees. 
 We combine the strengths of these two directions.
 
\begin{figure}[tbp]
    \centering
    \includegraphics[width=\linewidth]{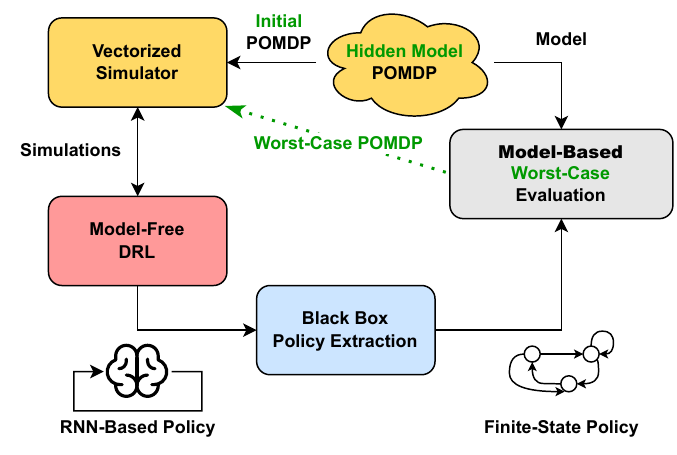}
    \caption{The overall idea of \toolname, our RL-based FSC extraction. The robust extensions are depicted in green.}
    \label{fig:rl-to-paynt-loop}
    \Description{Figure showing the \toolname robust neuro-symbolic loop.}
\end{figure}

\Cref{fig:rl-to-paynt-loop} depicts the three steps of \toolname: (1)~Training an RNN-based neural policy using DRL equipped with a vectorized simulator that provides highly efficient access to the known POMDP model. (2)~Extracting a stochastic finite-state controller (FSC) that mimics the behavior of the neural policy. 
 (3)~Model-based verification of the extracted FSC. 
  For extraction, we use two sampling-based methods: automata learning based on \textsc{Alergia}~\cite{DBLP:journals/ml/MaoCJNLN16} and a novel, scalable method that learns surrogate, self-interpretable networks, providing flexible control over FSC sizes. 
  Both methods are policy-architecture agnostic and treat policies as black boxes.
  We discuss prior work on using RNNs to find FSCs~\cite{pip-rnn,DBLP:journals/jair/Carr0T21} in the related work section. 
  
 We extend our approach towards \emph{robust} FSCs for \emph{hidden-model} POMDPs (HM-POMDPs)~\cite{hmpomdps,mepomdpnips25}, capturing different environments, i.e., sets of POMDPs with a shared action and observation space. 
A robustified version of \toolname searches for an FSC that achieves the best worst-case performance for a given HM-POMDP. 
 As depicted by the green parts of~\Cref{fig:rl-to-paynt-loop},
 for the candidate FSCs, we identify the worst-case POMDPs that iteratively enter the learning step.
 Similar to~\citet{hmpomdps}, a deductive verification technique~\cite{DBLP:conf/cav/AndriushchenkoC21} finds the worst-case POMDPs. 
 We sample trajectories from these worst-case POMDPs to train a robust neural policy and, subsequently, extract a robust FSC.

In summary, our main contributions are: 
\begin{enumerate}[topsep=2pt, itemsep=1.5pt, parsep=1.5pt, partopsep=0.5pt, leftmargin=20pt, labelsep=0.5em, labelindent=0pt, labelwidth=!]
\item \toolname, a scalable learning-based approach for computing high-quality FSCs for large POMDPs,
\item an extension of \toolname computing robust  
FSCs for large HM-POMDPs via iterative learning from worst-case POMDPs,
\item a scalable extraction method that approximates complex neural polices via stochastic FSCs. 
\end{enumerate}
In our experimental evaluation, 
we focus on planning problems with large state spaces that go beyond the reach of existing model-based approaches.
We demonstrate a clear advantage in terms of scalability when comparing \toolname to state-of-the-art approaches in both the single POMDP and HM-POMDP settings.

\section{Problem formulation}
\label{sec:problem-formulation}
In this section, we begin with notation and preliminaries, then formulate the central problems addressed in the paper.

The set of all distributions over a countable set $X$ is $\distr(X)$.
For $f \colon X \to \distr(Y)$, we may write $f(y \mid x)$ to denote $f(x)(y)$.
Below, we introduce 
(partially observable) \emph{Markov decision processes}~\cite{DBLP:books/wi/Puterman94}.

\subsection{MDPs, POMDPs and Policies}
\begin{definition}[MDP]
    A Markov decision process~(MDP) is a tuple $\mdp = \tuple{S, s_0, \actset, \trsfun, \rewfun}$, where $S$ represents a countable set of \emph{states}, $s_0 \in S$ is the \emph{initial state}, $\actset$ is a finite set of \emph{actions}, $\trsfun \colon S \times \actset \rightarrow \distr(S)$ is a \emph{transition function}, and $\rewfun \colon S\times \actset \rightarrow \mathbb{R} $ is a \emph{reward function}. 
\end{definition}
A \emph{Markov chain}~(MC) with rewards is an MDP without actions.
To handle environments that do not provide precise state information, MDPs are extended to model partially observable states.

\begin{definition}[POMDP]
    A partially observable Markov decision process~(POMDP)~\cite{kaelbling1998planning} is a tuple $\pomdp = (\mdp, \obsset, \obsfun)$, where $\mdp$ is an underlying MDP, $\obsset$ is a finite set of \emph{observations}, and $\obsfun \colon S \rightarrow \obsset$ is, w.l.o.g.~\cite{ChatterjeeCGK16}, a deterministic and state-based \emph{observation function}. 
\end{definition}

We focus on a general form of \emph{reachability} over an infinite horizon with rewards defined w.r.t.\ a set of \emph{target states} $G \subseteq S$ that are fully observable and absorbing~\cite{DBLP:conf/ijcai/HorakBC18}.
It encompasses (1) \emph{reachability}; maximizing reachability probabilities of target states $G$, and (2)~\emph{reachability reward}; maximizing cumulative reward before reaching target states $G$ with probability one.
For~(1), the reward function is one for states $g \in G$ and zero everywhere else.
For~(2), the states $g \in G$ have no reward and are reachable from all states.
Note that the reachability reward objective~(2) generalizes the discounted reward setting~\cite{DBLP:conf/ijcai/BonetG09}.
For a sequence of states $(s_0, s_1, \ldots, s_t)$ its observable fragment is the \emph{observation history} $o_t = (O(s_0), O(s_1), \ldots, O(s_t))$.
A \emph{trajectory} is a sequence containing observations and actions.
A \emph{policy} $\pi \in \Pi$ maps trajectories to distributions over actions.
We search for policies represented by finite-state controllers~\cite{DBLP:conf/nips/Hansen97,10.5555/2073796.2073844}.

\begin{definition}[FSC]
    A \emph{finite-state controller}~(FSC) is
    a tuple $\fsc = (N, n_0, \actfun, \memfun)$, where $N$ is a set of \emph{memory nodes} (states), $n_0 \in N$ is an \emph{initial node}, $\actfun \colon N \times Z \rightarrow \distr(A)$ is a (stochastic) \emph{action function}, and $\memfun \colon N \times Z \rightarrow \distr(N)$ is a (stochastic) \emph{memory update function}. 
\end{definition}
We employ stochastic Mealy-like FSCs~\cite{amato2010optimizing} that select actions based on memory and observations and thus require fewer memory nodes compared to Moore-like FSCs as used, e.g., in~\citet{you2025partially}.

The (uncountable) set of all (stochastic) FSCs for a POMDP $\pomdp$ is~$\FSC_{\pomdp}$.
A POMDP~$\pomdp$ together with an FSC~$\fsc$ constitutes the Markov chain~$\pomdp^{\fsc} = \tuple{S^{\fsc}, \tuple{s_0, n_0}, \trsfun^{\fsc}, \rewfun^{\fsc}}$ over the product state-space $S^{\fsc} = S \times N$, 
with, for all $\langle s, n\rangle \in S^{\fsc}$, using $z = O(s)$:
\begin{align*}
\trsfun^{\fsc}( \tuple{s',n'} \mid \tuple{s,n} ) &= \memfun(n' \mid n, z) \sum_{a\in\actset}  \actfun(a \mid n,z) \trsfun(s'\mid s,a), \\ 
\rewfun^{\fsc}( \tuple{s,n} ) &= \sum_{a\in\actset} \actfun(a \mid n,z) \rewfun(s,a).
\end{align*}
Consequently, the \emph{value-function} of an FSC $\fsc$ in a POMDP $\pomdp$ is the value-function $V^{\fsc} \colon S\times N \to \mathbb{R}$ of the MC $\pomdp^{\fsc}$, given by the fixed point of the following equation~\cite{10.5555/2073796.2073844}, again using $z = O(s)$:

\begin{equation*}
    V^{\fsc}(\tuple{s,n}) = \rewfun^{\fsc}(\tuple{s,n}) + 
    \sum_{\tuple{s',n'}
    } 
    \trsfun^{\fsc}( \tuple{s',n'} \mid \tuple{s,n} ) V^{\fsc}(\tuple{s',n'}).
\end{equation*}

Then, the \emph{value} $J^{\fsc} = V^{\fsc}(\tuple{s_0,n_0})$ of the FSC $\fsc$ is determined~ from the value function on the initial state of the MC $\pomdp^{\fsc}$.

\paragraph{Neural policies.}
Recurrent neural networks~(RNN), similarly to FSCs, represent state machines.
Nowadays, RNNs are a standard architecture implemented by gated recurrent units (GRU)~\cite{DBLP:journals/corr/ChoMBB14}, or long-short term memory (LSTM)~\cite{lstm_1997} layers that give DRL the capability of memory~\cite{hausknecht2015deep,DBLP:conf/icml/NiES22} and have been used to find FSCs for planning in (robust) POMDPs~\cite{DBLP:journals/jair/Carr0T21,pip-rnn}.
In contrast to FSCs, RNNs have an infinite space of memory, represented by the \emph{hidden state}
$\hiddenstate \in \mathbb{R}^d$, with size $d\in\mathbb{N}$ .
RNN policies typically consist of a parameterized (deterministic) memory update $\memfun_{\theta} \colon \mathbb{R}^d \times \obsset \to \mathbb{R}^d$ and a parameterized (stochastic) action function $\actfun_{\theta} \colon \mathbb{R}^d \to \distr(A)$.
Parameters $\theta$ are optimized by back-propagating the gradients of DRL losses through time~\cite{DBLP:journals/pieee/Werbos90,wierstra2007solving}. 
Just like an FSC, an RNN-based policy maps trajectories to actions and induces a value $J^{\rnn}$ representing the expected reward. 
It is, in general, intractable to compute it in closed form; thus, the value $J^{\rnn}$ is typically approximated empirically by simulating the policy $\rnn$ in the POMDP $\pomdp$.

\subsection{Hidden-model POMDPs}
Hidden-model POMDPs (HM-POMDPs) model a family of POMDPs that share the same state, observation, and action spaces but differ in their reward and transition functions. As such, HM-POMDPs capture uncertainty about the exact dynamics of the environments. 
Here, we largely follow~\citet{hmpomdps}.
\begin{definition}[HM-POMDP]
A \emph{hidden-model} POMDP~(HM-POMDP)~\cite{hmpomdps} is a tuple $\hmpomdp = \langle S, s_0,  \actset, \{\trsfun_i\}_{i\in I}, \{\rewfun_i\}_{i\in I}, \obsset, \obsfun\rangle$, where $I$ is a finite set of indices,  $S$, $s_0$, $\actset$, $\obsset$, and $\obsfun$ are defined as for standard POMDPs, 
$\{\trsfun_i\}_{i\in I}$ and $\{\rewfun_i\}_{i\in I}$ are indexed sets of transition and reward functions, respectively, such that, for every $i \in I$, $\pomdp_i = \langle S, s_0,  \actset, \trsfun_i, \rewfun_i, \obsset, \obsfun\rangle$ is a POMDP.
Consequently, an HM-POMDP $\hmpomdp$ describes a finite set (a family) of POMDPs $ \{\pomdp_i\}_{i \in I}$.
\end{definition}
Due to the shared action and observation spaces, all POMDPs in the HM-POMDP share the same set of FSCs $\FSC_{\hmpomdp}$.
Additionally, structural similarities, i.e., transitions with equivalent outcomes, can be merged in the construction of a \emph{quotient POMDP}~\cite{hmpomdps}.
Given a set of POMDPs, an FSC $\fsc$ induces a set (family) of MCs $\{M_i^{\fsc}\}_{i \in I}$~\cite{DBLP:conf/tacas/CeskaJJK19} and an associated set of value-functions $\{V_i^{\fsc}\}_{i \in I}$.
We are interested in the robust value, i.e., the worst-case value among the models. %

\begin{definition}[Robust value]
    \label{def:robval}
    Given an HM-POMDP $\hmpomdp$ and a policy $\fsc$, the \emph{robust value} is the worst-case value among the set:
    \begin{equation}
        \mathcal{J}^{\fsc} = \min_{i \in I} V_i^{\fsc}(\tuple{s_0, n_0}).
    \end{equation}
\end{definition}
As explained in \citet{hmpomdps}, the robust value can be found efficiently through deductive verification techniques~\cite{DBLP:conf/cav/AndriushchenkoC21}.

\subsection{Problem Statement}
\label{sec:problem}
Now, we are ready to define the main problems: maximizing reachability rewards (or probabilities) for either a single POMDP or an HM-POMDP.
First, we define our problem for individual POMDPs.
\begin{problem}
    \label{prob:pomdp}
    Given a POMDP $\pomdp$,
    find an FSC policy $\fsc$ that maximizes the value $V^{\fsc}(\tuple{s_0, n_0})$.
    Formally:
    \begin{equation}
        \fsc^* \in \argmax_{\fsc \in \FSC_{\pomdp}} V^{\fsc}(\tuple{s_0, n_0}).
    \end{equation}
\end{problem}
Similarly, we define the robust problem for HM-POMDPs:
\begin{problem}
    \label{prob:hmpomdp}
    Given an HM-POMDP $\hmpomdp$,
    find an FSC policy $\fsc$ that maximizes the robust value $\mathcal{J}^{\fsc}$ as in \Cref{def:robval}.
    Formally:
    \begin{equation}
        \fsc^* \in \argmax_{\fsc \in \FSC_{\hmpomdp}} \mathcal{J}^{\fsc} = \argmax_{\fsc \in \FSC_{\hmpomdp}} \min_{i \in I} {V}_i^{\fsc}(\tuple{s_0, n_0}).
    \end{equation}
\end{problem}

Note that Problem 2 resembles two-player zero-sum Stackelberg games~\cite{DBLP:journals/automatica/ZhengJL22}, which suggests that stochastic FSCs are necessary~\cite{hmpomdps}.
Similarly to POMDP planning with undiscounted infinite-horizon objectives, the decision variants of both \Cref{prob:pomdp} and \Cref{prob:hmpomdp} are undecidable in general~\cite{DBLP:conf/aaai/MadaniHC99}. 
Although there exist convergence guarantees for a variant of an infinite-horizon objective called \emph{Goal-POMDPs}~\cite{DBLP:conf/ijcai/HorakBC18}, which were recently also extended to a game setting~\cite{tomavsek2024iterative}, the practical performance of belief-based methods is limited~\cite{andriushchenko23search}, and the ideas cannot immediately be used for \Cref{prob:hmpomdp}. 
In this work, we aim to develop a scalable and sound algorithm to find the best possible FSC policy within a reasonable timeframe.

To that end, we compute (stochastic) FSCs using DRL, specifically using RNN-based agents.
In contrast to FSCs, the value of an RNN-based policy $\rnn$ cannot be found analytically, which is why we are ultimately interested in finding an FSC representation of~$\rnn$.

\section{On using DRL to solve POMDPs}

In the following, 
we motivate the use of DRL with RNNs for solving POMDPs, even when the model is fully known.

\subsection{Challenges of Solving POMDPs}
In POMDPs, it is well-known that acting near-optimally requires a succinct representation of (past) observation-action sequences~\cite{smallwood1973optimal}.
These sequences can be sufficiently compressed into \emph{beliefs}, i.e., probability distributions over the underlying unobserved states, and a near-optimal policy can be found by constructing and solving a \emph{belief-MDP}~\cite{kaelbling1998planning}.
Unfortunately, the complexity of updating a belief is quadratic in the size of the state-space~\cite{Spaan2012}, and the number of possible belief states grows exponentially in the horizon, rendering full exploration of belief-space intractable in general~\cite{DBLP:conf/aaai/MadaniHC99}.

\paragraph{Searching for FSCs.}
Exploration of FSCs offers an alternative method to belief-based policies 
with benefits in efficient verification and deployment.
One direction is to optimize the parameters of an FSC, where 
the size and structure of the memory must be established beforehand~\cite{amato2010optimizing, poupart2004bounded, junges2018finite}.  An alternative direction is to build the FSC incrementally through various symbolic search techniques ~\cite{kumar2015history,andriushchenko23search}. However, both directions suffer from an enormous design space.

\paragraph{Robust policies.}
For many situations, the assumption that a single POMDP model describes all cases is too strong.
Therefore, recent efforts focus on computing a robust policy for sets of POMDPs~\cite{DBLP:conf/aaai/Cubuktepe0JMST21,pip-rnn,hmpomdps,DBLP:conf/icml/Osogami15}. 
In a robust setting, the challenge is to find a single policy that works for (large) sets of POMDPs.
Thus, the scalability challenges and the importance of generalization are further increased, and existing approaches for POMDPs do not readily extend.

\subsection{Opportunities and Challenges of DRL}
To address the aforementioned challenges, we leverage model-free DRL, which does not rely on a model and instead optimizes over its simulations, enabling scaling to large state spaces.
DRL handles partial observability, by using 
RNNs 
as memory representations for policies~\cite{DBLP:conf/icml/NiES22,hausknecht2015deep,wierstra2007solving}, due to their capability of generalization~\cite{DBLP:conf/icml/NiES22} and learning sufficient statistics of observation sequences~\cite{DBLP:journals/tmlr/LambrechtsBE22}.

\paragraph{DRL for solving (HM-)POMDPs.}
In our setting, we are given a fully specified (HM-)POMDP model. 
As explained in the introduction, utilizing DRL to solve discrete (HM-)POMDPs presents its own challenges and remains an unsolved problem.
We must compute a policy in symbolic form, specifically an FSC, such that we can evaluate its performance analytically.
Thus, while we neglect the POMDP model in searching for the FSC, we still use it to provide guarantees on (i.e., verify) the policy.
We observe that using RNNs to search for FSCs for a given POMDP has been very successful~\cite{DBLP:journals/jair/Carr0T21,pip-rnn}.
However, so far, training the RNN has relied on model-based information, which suffers from the fact that (1) it does not scale and (2) relies on suboptimal approximations.
\toolname fills this gap.

\begin{algorithm}[tb]
\caption{Overview of \toolname and Robust \toolname}
\label{alg:lexpop}
\begin{algorithmic}[1]
\Function{TrainPolicy}{$\rnn$, $\{\pomdp_i\}_{i \in I'}$}
    \State $\text{VecSimulator} \gets \textsc{VecStorm}(\{\pomdp_i\}_{i \in I'})$ 
    \State $\rnn' \gets \textsc{DRL}(\rnn, \text{VecSimulator})$ \Comment{Using PPO}
    \State \Return $\rnn'$
\EndFunction


\Function{\toolname}{$\pomdp$: POMDP, $\rnn$ : Policy}
    \State $\rnn \gets \textsc{TrainPolicy}(\text{$\rnn$}, \{\pomdp\})$  
    \State $\fsc \gets \textsc{ExtractFSC}(\rnn)$ \Comment{Using Alergia or SIG}
    \State $J^{\fsc} \gets \textsc{Evaluate}(\fsc, \pomdp)$ \Comment{Using \textsc{Storm}}
    \State \Return $\fsc, J^{\fsc}$
\EndFunction

\Function{Robust\toolname}{$\hmpomdp$: HM-POMDP, $\rnn$ : Policy}
    \State $\pomdpbatch \gets \textsc{Initialize}()$  \Comment{$\pomdpbatch$ is a buffer of POMDPs}
    \While{ $\neg$ \textsc{TimeOut()} }
        \State $\rnn \gets \textsc{TrainPolicy}(\rnn, \pomdpbatch)$ \Comment{See App.~B}
        \State $\fsc \gets \textsc{ExtractFSC}(\rnn)$ \Comment{Using Alergia or SIG}
        \State $\pomdp, \mathcal{J}^{\fsc} \gets \textsc{worstPOMDP}(\fsc, \hmpomdp)$  \Comment{Using \paynt}
        \State $\pomdpbatch \gets \pomdpbatch \cup \{\pomdp\}$
    \EndWhile
    \State \Return $\fsc, \mathcal{J}^{\fsc}$
\EndFunction
\end{algorithmic}
\end{algorithm}
\section{Overview of our approach}

\Cref{alg:lexpop} provides an overview of \toolname.

\paragraph{(Robust) policy training}

\toolname is agnostic to the DRL algorithm used to train the neural policy. We chose PPO~\cite{ppo_schulmann}, due to its stable performance on our challenging POMDP benchmarks. For efficient sampling, we developed a vectorized simulation framework on top of the \storm model-checker, enabling parallel simulations across single or multiple POMDPs (see Line~2). In Line~3, our agent, represented by an RNN $\rnn$ with parameters $\theta$, is optimized on a (new) set of POMDPs through multiple training iterations. 
DRL for the considered (HM-)POMDPs poses significant technical challenges (cf.~\cite{challenges_rl}), including poorly defined and sparse rewards, partial observability, and dynamic action spaces. We provide a description of our DRL pipeline, including important design choices and alternative learning methods in~\Cref{appendix:rl_section}.

\paragraph{\toolname for \Cref{prob:pomdp}.}
\label{sec:controllers:for:pomdps}
Given a POMDP $\pomdp$, \textsc{\toolname} trains a neural policy $\rnn$ starting from a randomly initialized policy $\rnn$ (Line 6). The key step is the extraction of the FSC $\fsc$ from $\rnn$ (Line 7). We consider two methods discussed in Section~5. Finally, $\fsc$ is analytically evaluated by constructing and analyzing the Markov chain $\pomdp^{\fsc}$ in \storm~\cite{STORM}, yielding its value ${J}^{\fsc}$ (Line~8).

\paragraph{\toolname for \Cref{prob:hmpomdp}.}
Given an HM-POMDP $\hmpomdp$, \textsc{Robust\toolname} iteratively trains a robust neural policy and consequently extracts a robust FSC policy that performs well on the given set of POMDPs.
\textsc{Robust\toolname} maintains a candidate neural policy $\rnn$ and its finite-state approximation $\fsc$. Every candidate $\fsc$ is associated with its worst-case POMDP $\pomdp$ from $\hmpomdp$. These worst-case POMDPs are iteratively stored in a buffer $\pomdpbatch$. We start with a randomly initialized $\rnn$ and a randomly selected $\pomdp$ (Line 11). In every iteration, we improve $\rnn$ through robust policy training, using the trajectories collected from all POMDPs stored in the buffer $\pomdpbatch$ (Line~13). \Cref{appendix:robust_learning} provides more details. Then, we extract an improved $\fsc$ (Line 14) using the same extraction methods as in the single-POMDP setting (see~\Cref{sec:fsc-extraction}). Finally, we analytically compute the robust value $\mathcal{J}^{\fsc}$ associated with $\fsc$ and the corresponding worst-case POMDP $\pomdp$ (Line~15) which we add to the buffer $\pomdpbatch$ (Line~16). We employ \paynt~\cite{DBLP:conf/cav/AndriushchenkoC21} to efficiently search through the family of Markov chains $\{\pomdp_i^{\fsc}\}_{i\in I}$ induced by $\hmpomdp$ and~$\fsc$.

\section{Policy Extraction}
\label{sec:fsc-extraction}

The extraction of FSCs from RNN-based policies connects model-free policy learning and model-based policy verification. 
It relies on two insights: 1) for RNN policies, there often exists a smaller finite-state representation~\cite {DBLP:conf/iclr/KoulFG19} and 2) for challenging POMDPs, there often exists a small FSC achieving a close-to-optimal value~\cite{andriushchenko23search}.

To retrieve an FSC, we propose two extraction methods with a common working principle:
Provided an RNN policy $\rnn$, we synthesize an FSC policy $\fsc$ whose decisions match $\rnn$ on sampled training trajectories as closely as possible. Both methods learn stochastic FSCs from a sampled dataset of trajectories $\mathcal{D}$ from 
a subset of POMDPs $\{M_i\}_{i \in I'}$ using policy $\rnn$ trained on that subset.
The data set contains sequences of tuples $(\obs_t, a_t)$ of observations and actions selected by $\rnn$ at time $t$.
The first extraction method is a direct application of the automata learning algorithm \textsc{Alergia}~\cite{DBLP:conf/icgi/CarrascoO94}, whereas the second method introduces a novel RNN training approach with a hidden-state space~encoding.

\subsection{Automata Learning}
We learn stochastic Mealy machines~\cite{DBLP:journals/sosym/MuskardinTAP24} using a variant of \textsc{Alergia} proposed for MDPs~\cite{DBLP:journals/ml/MaoCJNLN16} and implemented in AALpy~\cite{Musardin2022}. Therefore, we slightly adjust the FSC definition to  $\fsc = (N, n_0, \actmemfun)$, combining the memory update and action function into $\actmemfun : N \times Z \rightarrow \distr(N \times A)$.

 Given the dataset $\mathcal{D}$ of trajectories, \textsc{Alergia} first transforms them into a prefix tree by merging common prefixes. Then, it merges tree nodes having compatible subtrees, i.e., nodes whose observed future behavior is similar. Compatibility checks recursively determine if the action distributions conditioned on observation-action histories are not statistically different. Therefore, merging compatible nodes creates an FSC memory structure that mimics $\rnn$. After processing all possible merges, \textsc{Alergia} estimates action and update probability distributions. Instead of commonly used compatibility checks based on Hoeffding bounds~\cite{DBLP:conf/icgi/CarrascoO94,DBLP:journals/ml/MaoCJNLN16}, we use $\chi^2$ homogeneity tests.
This better distinguishes memory nodes, at the cost of slightly larger FSCs and, thus, longer verification times.
In the limit, \textsc{Alergia}~\cite{DBLP:journals/ml/MaoCJNLN16} converges to an FSC that is isomorphic to the FSC underlying the RNN policy, if such underlying FSC exists.

\subsection{Self-Interpretable Networks}
As automata learning typically struggles with large action and observation spaces, we additionally
propose an approach that learns interpretable neural surrogate models.  
\newcommand{\fscnet}{\pi_{\text{FSC}_{\phi}}}
The core idea is to approximate the behavior of the policy $\rnn$ using a neural network $\fscnet$, which is explicitly designed to emulate an FSC. The idea of using a discrete memory structure is inspired by \emph{quantized bottleneck layers} proposed by~\citet{DBLP:conf/iclr/KoulFG19}, which employs a hyperbolic tangent ($\tanh$) layer followed by rounding to quantize the RNN feedback. Training with quantization requires a straight-through gradient estimator.
 Our extraction method, \emph{Self-Interpretable Gumbel Softmax Network}~(SIG), mitigates three significant limitations arising from this design:
 a)~an exponential blow-up of the number of memory nodes caused by the flat quantization of the activation values; b)~the lack of support for stochastic memory updates; and c)~the bias of the straight-through gradient estimates.
To combat these issues, we propose a bottleneck architecture that directly models the transition distribution to one of $\statesn$ memory states.

\begin{figure}
    \centering
    \includegraphics[width=0.9\linewidth]{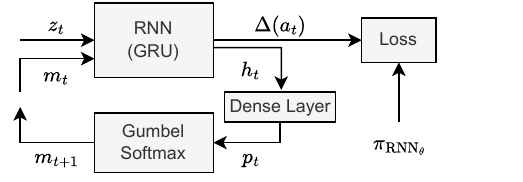}
    \caption{High-level architecture of the self-interpretable Gumbel softmax network.}
    \label{fig:gumbel_si}
    \Description{Figure showing our architecture of self-interpretable Gumbel softmax relying on bottlenecked GRU feedback.}
\end{figure}
\paragraph{Architecture and Training.}
\Cref{fig:gumbel_si} depicts the architecture for training the surrogate policy $\fscnet$. 
The recurrent part transforms hidden states $\hiddenstate_t \in \mathbb{R}^d$ into a space of vectors $m_{t+1} \in [0,1]^\statesn$ that resembles categorical distributions over one-hot encoded memory nodes. The size $\statesn$ is a parameter of extraction.  
The forward pass first computes the vector $p_t \in \mathbb{R}^\statesn$ of non-normalized log-probabilities of the memory update. While the distilled FSC directly samples $m_{t+1} \sim \softmax(p_t)$ to update the memory state, this approach is not differentiable. Therefore, during training, we use a Gumbel Softmax layer~\cite{jang2017categorical} which is designed to sample $m_{t+1}$ in a differentiable way. We pay the cost that $m_{t+1}$ is no longer a one-hot vector, though its distribution is close to $\softmax(p_t)$ (in the Wasserstein distance); the distance is controlled by a temperature parameter and is reduced during training.
To train the SIG network, we minimize the empirical cross-entropy between the actions from the dataset $\mathcal{D}$ and the computed actions. 

\paragraph{FSC Inference.}
To construct an FSC, we infer the actions and memory for each combination of memory node and input observation~\cite{pip-rnn}. 
Given a trained SIG with a bottleneck layer of size $\statesn$, we create a placeholder FSC $\fsc$ with memory nodes $N$.
Then, for every node $n \in N$ and observation $\obs \in \obsset$, we perform these steps:
\begin{enumerate}
    \item Create a one-hot encoding $m \in \{0,1\}^{|N|}$ of $n$.
    \item Process $\obs$ and $m$ through the RNN to yield an action distribution $\sigma_a \in \distr(A)$ and hidden state $\hiddenstate
    \in\mathbb{R}^d$.
    \item Process $\hiddenstate$ with the Gumbel encoder to get a distribution $\sigma_{n'} \in \distr(N)$ over the next memory node $n' \in N$.
    \item Add $(n,z) \mapsto \sigma_a$ to $\actfun$ and $(n,z) \mapsto \sigma_{n'}$ to $\memfun$ of the FSC $\fsc$.
\end{enumerate}
Then, the above yields a fully specified FSC $\fsc$. We prune nodes that are not reachable from the initial observation-memory tuple.
While SIG does not provide convergence guarantees like \textsc{Alergia}, we find that SIG works well and often even better in practice.

\section{Experiments}
\label{sec:exp}

\newcommand{\best}[1]{\textbf{#1}}

We derive two main research questions from the problems formulated in \Cref{sec:problem}, and further refine them into sub-questions, considering the individual components of \toolname:

\noindent
\textbf{RQ1.} \emph{Can \toolname construct FSCs with a higher value than FSCs synthesized by a state-of-the-art POMDP solver \saynt~\cite{andriushchenko23search}?}

\begin{enumerate}[label=\textbf{RQ1\alph*}]
    \item Can the PPO algorithm train RNN-based neural policies that empirically outperform the value of FSCs from \saynt?
    \item Does the extraction of FSCs from the neural policies preserve the value?
    \item Does the self-interpretable extraction improve the fidelity over stochastic automata learning using \textsc{Alergia}~\cite{DBLP:journals/ml/MaoCJNLN16}?
\end{enumerate}

\noindent
\textbf{RQ2.} \emph{For the given HM-POMDPs, can \toolname construct robust FSC policies with a higher robust value than the FSCs produced by the state-of-the-art model-based algorithm \rfpg~\cite{hmpomdps}?}

\begin{enumerate}[label=\textbf{RQ2\alph*}]
    \item Does the FSC extraction preserve the robust value of the robustified neural policy on the given HM-POMDPs?
    \item Can the robustified PPO algorithm, combined with FSC extraction, outperform the FSCs produced by \rfpg?
    \item Is the worst-case POMDP selection important for solving HM-POMDPs? 
    
\end{enumerate}

\begin{table*}[th]
     \centering

    \caption{Results for the single POMDP setting using 30-minute timeout. In all models, the goal is to maximize the value. For \toolname, we report the interquartile mean (IQM) of the empirical values of the neural policies and the IQM of the extracted FSCs, along with their average sizes, using the two extraction methods. The values in brackets show the interquartile range (IQR). We highlight the best value(s) with statistical significance, excluding the empirical value for \toolname.
    }
    \begin{tabular}{c||ccc||c||c||cc||cc}
    \toprule
    & \multicolumn{3}{c||}{Dimensions} & \saynt & \toolname (RNN) & \multicolumn{2}{c||}{\toolname (Alergia)} & \multicolumn{2}{c}{\toolname (SIG)} \\
    Model & $|S|$ & |Z| & |A| & Value & IQM (IQR) & IQM (IQR) & FSC Size & IQM (IQR) & FSC Size \\
    \midrule
    maze-10 & 4.8k & 22 & 4 & \best{8.59} & 8.86 (0.03) & \best{8.52} (0.10) & 20.0 & \best{8.54} (0.47) & 3.0 \\ 
    rocks-16 & 11k & 2.7k & 10 & \best{-36.91} & -48.17 (1.43) & -51.49 (1.41) & 6.0 & -48.70 (1.28) & 3.0 \\
    network-3-8-20 & 17k& 2.2k & 5 & -10.45 & -7.36 (0.27) & -8.07 (0.38) & 5.0 & \best{-7.70} (0.15) & 3.0 \\
    network-5-10-8 & 117k & 3.7k & 8 & -16.12 & -12.56 (0.20) & -13.76 (0.92) & 11.0 & \best{-12.71} (0.14)  & 3.0 \\
    intercept-16 & 130k & 21k & 6 & 0.80 & 1.00 (0.00) & 0.98 (0.01) & 6.0 & \best{0.99} (0.01) & 3.0 \\
    evade-n17 & 314k & 158k & 7 & 0.58 & 0.85 (0.02) & \best{0.85} (0.01) & 4.0 & \best{0.84} (0.01) & 3.0 \\
    drone-2-8-1 & 520k & 889 & 6 & 0.40 & 0.61 (0.01) & \best{0.58} (0.01) & 5.0 & \best{0.58} (0.01) & 3.0 \\
    \bottomrule
    \end{tabular}

    \label{tab:single_pomdp_results}
\end{table*}

\begin{table*}[tbh]
\caption{Results for the HM-POMDP setting using 1-hour timeout. The goal is to maximize the robust value.
We report the interquartile mean and the best robust value achieved by FSCs produced by \rfpg and by the two variants of \toolname, together with the average reachable memory of the extracted FSCs. The highlighted value(s) are the best with statistical significance.
}
\begin{tabular}{c||cccc||cc||ccc||ccc}
\toprule
& \multicolumn{4}{c||}{Dimensions} & \multicolumn{2}{c||}{\rfpg} & \multicolumn{3}{c||}{\toolname (Alergia)} & \multicolumn{3}{c}{\toolname (SIG)} \\
Model & $|S|$ & $|I|$ & $|\obsset|$ & |A| & IQM & Best  & IQM & Best & FSC Size &  IQM & Best & FSC Size\\
\midrule
rover & 86 & 6000 & 18 & 5 & \best{299.24} & 299.40 & 290.17 & 296.14 & 7.70 & 294.68 & 297.65 & 2.8\\
obstacles-8-5 & 380 & 12000 & 25 & 9 & \best{-205.05} & -193.93 & -225.98  & -212.46 & 2.50 & -218.29 & -207.63 & 2.9 \\
network & 4k & 140 & 20 & 6 & 3.51 & 3.78 & \best{3.76} & 3.80 & 4.30 & \best{3.78} & 3.82 & 3.0 \\
avoid & 13k & 1600 & 11 & 9 & -174.23 & -139.84 & \best{-161.01} & -145.87 & 6.60 & -163.15 & -140.88 & 3.0 \\
drone-2-6-1 & 44k & 3200 & 432 & 7 & 0.01 & 0.04 & \best{0.59} & 0.62 & 7.50 & \best{0.59} & 0.62 & 10.0 \\
avoid-large & 45k & 7056 &  11 & 9 & -228.42 & -166.14 & \best{-172.80} & -154.02 & 4.40 & \best{-170.53} & -141.63 & 8.8 \\
moving\_obstacles & 262k & 400 & 14 & 8 & -2113.79 & -2036.71 & -158.98 & -121.98 & 10.10 & \best{-86.86} & -76.45 & 5.2 \\
\bottomrule
\end{tabular}
\label{tab:robust_results}
\end{table*}

\paragraph{Baseline Selection}
RQ1 concerns the experimental comparison of DRL-based FSC synthesis and offline model-based planning. As a baseline, we selected one of the state-of-the-art approaches for offline POMDP planning, \saynt~\cite{andriushchenko23search}: it has publicly available code, supports both reachability and undiscounted reward specifications, and to our knowledge, works with the largest models. 
We also considered recent work by~\citet{uai24}, which extends the heuristic search value iteration algorithm~\cite{smith2004heuristic} to reachability specifications, outperforming SARSOP~\cite{kurniawati2008sarsop}. 
Our preliminary experiments show that its implementation cannot handle our large reachability models, which are about two orders of magnitude larger than those in~\cite{uai24}.

Unfortunately, we were unable to run the approach described in~\citet{DBLP:journals/jair/Carr0T21}, which also employs RNNs to extract FSCs. 
To our knowledge, the recently proposed \rfpg algorithm~\cite{hmpomdps} is the only available method that allows a direct comparison on HM-POMDPs within RQ2.
The other approaches discussed in the related work, see~\Cref{sec:related}, consider different formulations.

\paragraph{Benchmark Sets}
The experimental evaluation is based on two benchmark sets. 
For RQ1, we use both original and enlarged POMDP models in the PRISM format~\cite{KNP11} from the~\saynt repository~\cite{ andriushchenko23search}. 
For RQ2, we consider HM-POMDPs used to evaluate~\rfpg~\cite{hmpomdps}, excluding the DPM model with no observation features. Moreover, we constructed new HM-POMDPs by extending models used for RQ1 and by scaling up the original models in~\cite{hmpomdps}.
\Cref{appendix:stats_section} provides more details on the benchmarks.
The model statistics in~\Cref{tab:single_pomdp_results,tab:robust_results} are: $|S|$ is the maximum number of reachable states across all POMDPs, and $|I|$, $|Z|$, $|A|$ are the number of POMDPs, observations, and actions, respectively.  
The goal is to maximize the reachability probability or the reachability reward. Our benchmarks are challenging for DRL due to sparse rewards and dynamic discrete action spaces as discussed in~\Cref{appendix:dynamic_action_space}.

\paragraph{Experimental Setting}
We run experiments on a Ryzen~7~7840U processor with 32 GB of RAM. \toolname uses the PPO~\cite{ppo_schulmann} implementation in the TensorFlow Agents framework~\cite{TFAgents}, and a vectorized simulator implemented in the JAX framework~\cite{jax2018github}.
Verification of FSCs and worst-case POMDP synthesis are performed using the PAYNT~\cite{DBLP:conf/cav/AndriushchenkoC21} toolkit based on the Storm model checker~\cite{STORM}. Neural policies are evaluated using $512$ independent simulations in a given 
POMDP, performed in parallel using the vectorized simulator.
We truncate simulations after a fixed number of steps (typically 600) to have at least 512 finished episodes. For the single-POMDP setting, such an evaluation provides us with statistically reliable estimates of the value. In the case of HM-POMDPs, the independent simulations are performed on a multiset of POMDPs in the current POMDP buffer (see~\Cref{alg:lexpop}). \Cref{appendix:evaluation} provides more details.
For automata learning, we use \textsc{Alergia}~\cite{DBLP:journals/ml/MaoCJNLN16} implemented in AALpy~\cite{Musardin2022}. The \toolname implementation is publicly available in~\cite{implementation}.


\begin{figure*}[th]
    \centering
    \begin{subfigure}{.24\textwidth}
    \centering
    \includegraphics[width=.95\linewidth]{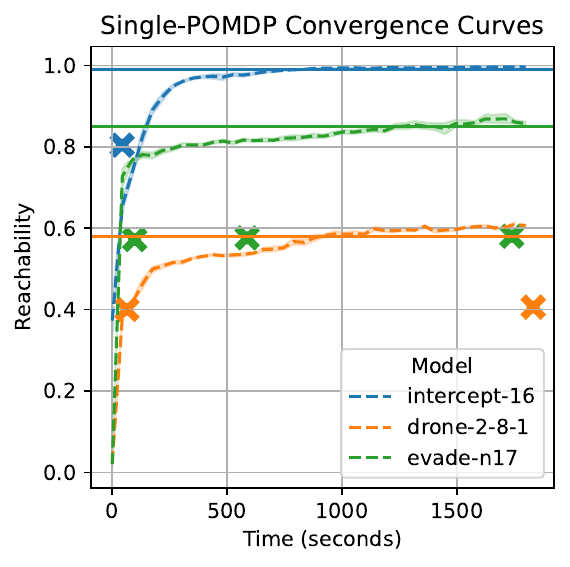}
    \caption{Large POMDPs}
    \end{subfigure}%
    \begin{subfigure}{.24\textwidth}
        \centering
        \includegraphics[width=.95\linewidth]{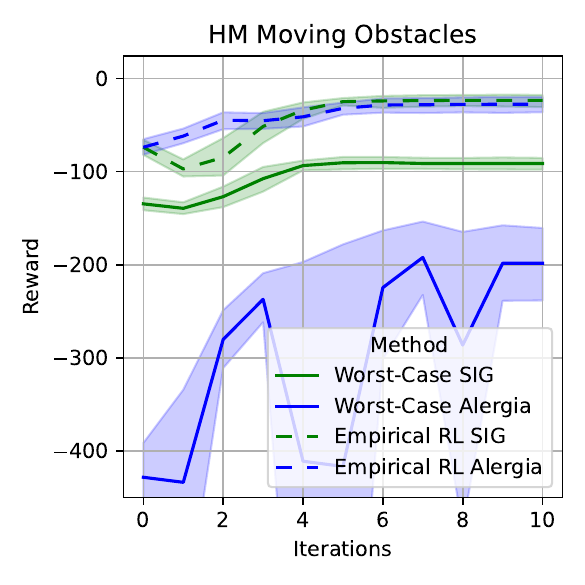}
        \caption{Robust Moving}
    \end{subfigure}
    \begin{subfigure}{.24\textwidth}
        \centering
        \includegraphics[width=.95\linewidth]{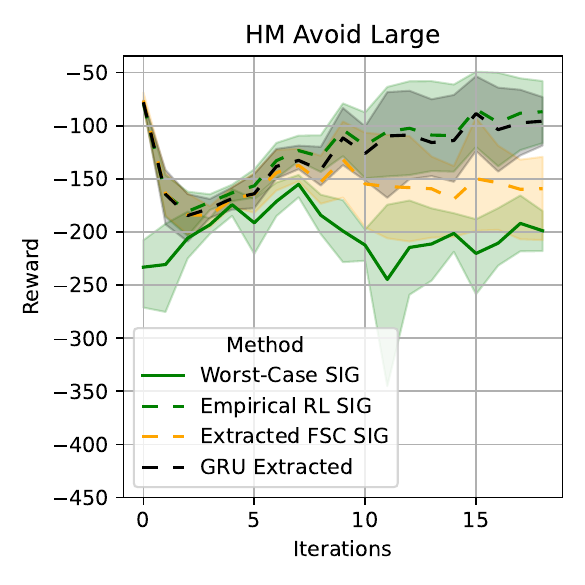}
        \caption{Robust Avoid-Large}
    \end{subfigure}
     \begin{subfigure}{.24\textwidth}
        \centering
        \includegraphics[width=.95\linewidth]{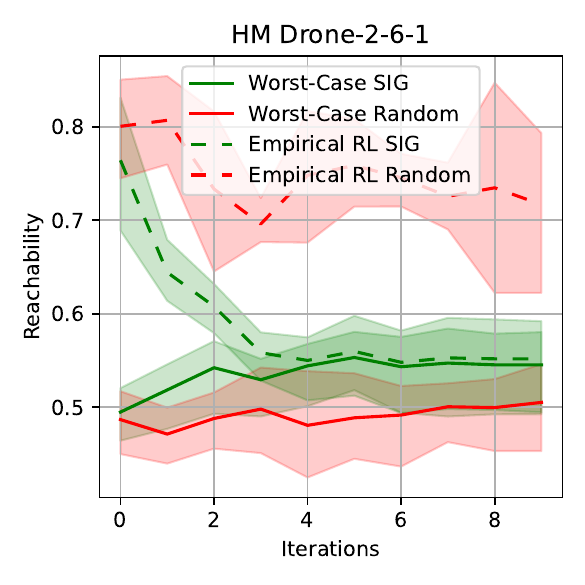}
        \caption{Robust Drone}
    \end{subfigure}
    \caption{Selected convergence curves showing average values and IQR (if applicable): (a) Large single-POMDP benchmarks: Empirical probabilities of neural policies (dashed lines), and values of the FSC produced by \saynt (crosses) and extracted from the final neural policy (solid lines).
    (b,c) Performance loss of the FSC extraction: Empirical value of the neural policy on the set of POMDPs used for training (dashed lines) and robust values of the extracted FSCs in the HM-POMDPs (solid lines). 
    Additionally, in (c), the orange line shows the empirical value of the extracted FSCs on the set of POMDPs used for training, and the black line shows the empirical value achieved by the same self-interpretable network without the quantized layers on the same set of POMDPs.
    (d) Impact of the worst-case POMDP selection in HM-POMDPs: Lines as in (b).  Green lines show the results using the worst-case POMDP selection, red lines show the results for random selection. Additional convergence curves are in~\Cref{appendix:extended_results}.} 
    \label{fig:curves}
    \Description{Convergence curves for both standard and robust \toolname implementations.}
\end{figure*}

\subsection*{RQ1: POMDP Planning}
\paragraph{Setting} We train each neural policy for 4000 iterations on every model. Following DRL training, we extract FSCs with both extraction methods.
The training takes about 23 minutes, while FSC extraction takes between 2 and 8 minutes. 
For a fair comparison, we configured \saynt with a 30-minute timeout on the FSC synthesis. 
\saynt is deterministic and therefore runs only once.
For \toolname, we repeat each run 10 times with fixed random seeds and report the interquartile mean~(IQM) and the interquartile range~(IQR).

{\textbf{Main results for RQ1}.} \Cref{tab:single_pomdp_results} 
 shows the FSC values achieved by \saynt, the average empirical value of the learned neural policy, and the average value of the FSC extracted by \toolname from the neural policies using the two variants of the extraction method. 
The table also reports the average sizes of the FSCs produced by \toolname, see~\Cref{appendix:architectures} for more details.

{\textbf{RQ1a}.} 
\Cref{tab:single_pomdp_results} shows that the learned neural policies provide better empirical values than the verified value of the FSCs produced by \saynt when the model gets larger, and comparable ones for small models, e.g., \texttt{rocks-16}, where \saynt benefits from the full knowledge of the model.
 \Cref{fig:curves}~(a) provides a more in-depth performance assessment on large models. The plots compare the empirical values of the reachability probability achieved by neural policies (dashed lines) with the FSC values synthesized by \saynt (crosses). We observe that for large models, \saynt relies primarily on an initial memory-less policy and cannot improve over time, whereas PPO continuously improves the policy. The main bottleneck of \saynt is the size of the considered FSCs, combined with the large number of beliefs that must be explored.
\textbf{Summary: For a large majority of models, PPO learns policies with better empirical values than verified controllers produced by \saynt.}

{\textbf{RQ1b}.} 
Next, we examine whether \toolname can extract an FSC from the given neural policy with sufficient fidelity. 
The results in~\Cref{tab:single_pomdp_results} generally show that the extracted FSC policies perform similarly to the original neural policies and thus significantly outperform FSCs produced by~\saynt except for the \texttt{rocks-16} model. 
This holds for both extraction approaches, as confirmed by a Sign test, which yields a p-value of approximately 0.001 for all models except for the mentioned \texttt{rocks-16} and \texttt{maze-10}, as shown in~\Cref{appendix:stats_section}. Our results also confirm that we can extract small FSCs that achieve similar performance. Another benefit compared to \saynt controllers is the simpler structure of our extracted FSCs. 
In contrast, \saynt typically uses a more complex structure to represent policies that combine FSCs with belief-based randomized policies~\cite{andriushchenko23search}. 
\textbf{Summary: Although abstraction of FSCs incurs a slight degradation of the achieved values, \toolname extracts verified FSCs that outperform the FSCs produced by \saynt.}

{\textbf{RQ1c}.} \Cref{tab:single_pomdp_results} shows that the SIG extraction provides an advantage over \textsc{Alergia}~\cite{DBLP:journals/ml/MaoCJNLN16} for the \texttt{rocks} and \texttt{network} models. 
In these cases, it yields FSCs with higher values that are close to the empirical values of the neural policies, whereas for the rest of the models, the results are mostly similar. However, the main advantage of the SIG extraction is its flexibility. The size of the final FSC learned by \textsc{Alergia} and the extraction time cannot be easily controlled -- they mainly depend on the input data. In contrast, self-interpretation can provide the best FSC with a user-defined number of memory nodes. This significantly affects verification time, which increases with the controller's size. In the worst case, we observed a six-fold increase in verification time when using \textsc{Alergia}.
\textbf{Summary: The proposed SIG extraction produces smaller FSCs than automata learning while preserving the performance.}

\subsection*{RQ2: Robust POMDP Planning}
\paragraph{Setting}
All experiments run with a 1-hour timeout and 10 
different seeds. The number of outer iterations (Line 12 of \Cref{alg:lexpop}) varies between each model and extraction method. 
Outer iterations of the algorithm might use different sets of POMDPs when using a different extraction algorithm, since the extracted FSCs affect the selection of worst-case POMDPs.

{\textbf{Main results}.} \Cref{tab:robust_results} 
includes the main results for RQ2. It shows the average and the best robust value achieved by FSCs produced by \rfpg and by the two variants of \toolname. It also reports the average size of the extracted FSCs (see~\Cref{appendix:architectures}). In the case of \rfpg, the size of the FSCs is predetermined by a heuristic and varies across observations, with an upper bound of 4~\cite{hmpomdps}.

{\textbf{RQ2a}.} 
The extraction of FSCs from robustified neural policies is generally more challenging, since
the hidden states have to handle the partial observability as well as the uncertainty in the environment dynamics. As a result, the hidden state space can be significantly more complex. \Cref{fig:curves}~(b) shows a nontrivial performance gap between the neural policy (dashed lines) evaluated on the set of POMDPs used for training and the robust value of the extracted FSC, evaluated on the whole HM-POMDP, in the \texttt{moving-obstacle} model. Moreover, we observe instability in the extraction using \textsc{Alergia}, which results in a less robust FSC compared to the FSC extracted using the SIG method. 

\Cref{fig:curves}~(c) shows the results for \texttt{avoid-large} model where the performance gap increases as the neural policy improves.
To investigate this phenomenon, we consider a GRU~\cite{DBLP:journals/corr/ChoMBB14} modification of self-interpretable networks, see~\Cref{appendix:architectures}, where we do not use the quantized layers and rely on continuous values of the GRU feedback. 
The performance of the modified network (black line) aligns with that of the original neural policy. This shows that the performance gap of the extracted FSCs is caused by the imposed memory limitation, not by the training process.  
\textbf{Summary: For complex HM-POMDP benchmarks, the FSC extraction is challenging due to larger memory requirements.
The SIG method provides better and more stable extraction.} 

{\textbf{RQ2b}.} 
We now compare the performance of \toolname with \rfpg. \Cref{tab:robust_results} shows similar trends as in RQ1: \toolname achieves worse results than \rfpg on small models, but significantly outperforms \rfpg on larger models. The statistical significance is confirmed with a Brunner Munzel test, as shown in~\Cref{appendix:extended_results}. 
The poor performance of \rfpg on large models is caused by the expensive parameter updates, which rely on gradient computation through the complete state space. This is exacerbated in the case of the \texttt{drone} model and the larger version of the \texttt{moving\_obstacles} model. For these models, we needed to reduce the size of FSCs synthesized by \rfpg to 1, since the gradient computation was extremely expensive and could not finish the first iteration. 
\textbf{Summary: \toolname scales considerably better than \rfpg and thus produces significantly better FSCs on large models.}

{\textbf{RQ2c}.} Our results confirm the observations from \rfpg~\cite{hmpomdps}: For HM-POMDPs describing a large set of POMDPs, a random selection of POMDPs for training robust policies is not sufficient. 
\Cref{fig:curves}~(d) illustrates this phenomenon on the \texttt{drone-2-6-1} model. Using randomly selected POMDPs makes the robust training unstable: the red dashed line shows the empirical value of the neural policy evaluated on the set of POMDPs used for training. More importantly, there is a very large performance gap between the neural policy and the robust value of the extracted FSC (the red solid line), regardless of the extraction method. On the other hand, using the worst-case POMDPs (the green lines) leads to more stable learning, a smaller performance gap, and consequently to an FSC having a considerably better robust value. 
\textbf{Summary: The worst-case POMDP selection is crucial for solving HM-POMDPs.}

\section{Related Work}
\label{sec:related}

\paragraph{POMDP solvers}
Existing work includes (monolithic) mixed integer linear program formulation for FSC optimization~\cite{DBLP:conf/aaai/AmatoBZ10,kumar2015history,DBLP:journals/sttt/WintererWBJ24}, game-based abstraction techniques~\cite{DBLP:journals/tac/WintererJWJTKB21}, or the approximate analysis of belief space~\cite{smith2004heuristic} implemented e.g. in SARSOP~\cite{kurniawati2008sarsop} that has been recently extended to unbounded reachability objectives~\cite{uai24}.
\saynt~\cite{andriushchenko23search} combines an approximate exploration of belief states and symbolic search techniques over FSCs.
\citet{DBLP:journals/jair/Carr0T21} proposed for the first time to extract FSCs from RNNs to compute verified FSC policies for POMDPs.
Yet, this work employs a simplistic, non-vectorized approach that optimizes MDP policies to train the RNN and an extraction method that lacks sufficient flexibility in FSC size. 
Note that, despite using the RNN policy, this approach does not employ DRL to optimize the RNNs.
  Overall, our approach significantly improves scalability by enabling the computation of FSC policies for larger POMDP models.

\paragraph{Robust FSCs for POMDPs}
\citet{hmpomdps} introduce HM-POMDPs that describe finitely many variations of a POMDP model, and propose the \rfpg algorithm that uses model-based subgradient ascent to iteratively optimize FSCs over worst-case POMDPs. 
Robust POMDPs~\cite{DBLP:conf/icml/Osogami15,bovy2024imprecise,rpomdpnips25}
represent infinite sets of POMDPs, encapsulating HM-POMDPs, where FSCs have been optimized through
 convex programming~\cite{DBLP:conf/aaai/Cubuktepe0JMST21,DBLP:conf/ijcai/Suilen0CT20}.
\citet{pip-rnn} introduce pessimistic iterative planning using RNNs for robust POMDPs. 
While conceptually similar, they rely on model-based approximations to train the RNNs instead of DRL, hindering scalability and optimality.

\paragraph{Finite-state model extraction from RNNs.}
The extraction of FSCs originates from
extracting finite-state models from RNNs, which has been explored already in the 1990s~\cite{DBLP:journals/neco/ZengGS93,DBLP:journals/nn/OmlinG96} and has recently gained attention.
There are two major approaches:
automata learning ~\cite{DBLP:conf/icml/WeissGY18,DBLP:conf/cdmake/MayrY18,DBLP:conf/ifm/MuskardinAPT22} from the trajectories sampled from the RNN,
and quantization of hidden states of RNNs~\cite{DBLP:conf/iclr/KoulFG19,pip-rnn}.
In line with~\citet{DBLP:conf/iclr/KoulFG19},~\citet{carr2019counterexample,DBLP:conf/ijcai/CarrJT20,DBLP:journals/jair/Carr0T21},  and~\citet{pip-rnn}, we also experiment with ways to extract an FSC directly from the RNN. 
Compared to these, our proposed architecture enhances control over memory size and stabilizes learning by eliminating the straight-through gradient estimate by reparametrization using Gumbel softmax distribution~\cite{jang2017categorical}.
Though similar in principle to the work of~\citet{DBLP:journals/sosym/AichernigKMPT24}, our architecture improves upon the prior work
by enabling stochastic action and memory state-updates, and achieving faster convergence.

\paragraph{Robust DRL} 

Various DRL methods address model uncertainty. The most similar approaches to ours are task-robust model-agnostic meta-learning~\cite{maml}, which combines meta-learning with worst-case optimization, and robust adversarial learning~\cite{rarl, gleave_2021, huang_2017, DBLP:conf/iclr/ZhangCBH21}, which frames learning as a stochastic game between a protagonist and adversary. Since we aim to extract a universal FSC, meta-learning is of limited applicability. In relation to~\cite{rarl}, our approach can be seen as using \paynt as an adversary, which is possible thanks to explicit knowledge of the model family. Other notable methods include model-free worst-case estimation~\cite{DBLP:conf/icml/GadotWKLM24}, training-time perturbations of states, actions, or observations~\cite{liu_2023}, and reward shaping for worst-case optimization, e.g., via state entropy~\cite{massiani2024viability}.

\section{Conclusion}

We proposed \toolname, a learning-based approach that constructs FSCs for a given POMDP or HM-POMDP. 
First, \toolname trains an RNN-based neural policy using DRL, which allows us to scale to large model sizes. 
Second, \toolname extracts FSCs from the neural policies using automata learning as well as a novel extraction technique. 
The performance of such FSCs can be formally verified on large POMDPs and for HM-POMDPs, worst-case POMDPs can be effectively computed to iteratively construct robust FSCs.
Our experiments demonstrate that \toolname scales up to large (HM-)POMDPs with state spaces that were previously infeasible. We plan to improve the robust planning loop by extracting programmatic policies~\cite{pmlr-v80-verma18a}. Furthermore, we plan to scale the loop even further by extending SIG towards model learning.

\section*{Acknowledgments}
\setlength{\intextsep}{0pt}
\begin{wrapfigure}{l}{.35cm}
\includegraphics[width=.9cm]{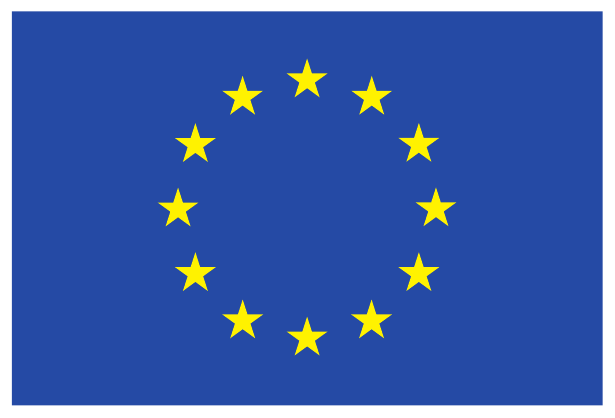}
\end{wrapfigure}
 This work has been executed under the project VASSAL: “Verification and Analysis for Safety and Security of Applications in Life” funded by the European Union under Horizon Europe WIDERA Coordination and Support Action/Grant Agreement No.101160022. Additionally, this work has been partially funded by the Czech Science Foundation grant GA23-06963S (VESCAA), the Vienna Science and Technology Fund (WWTF) project \href{https://taiger.logic.at}{10.47379/ICT22-023}, and the ERC Starting Grant DEUCE (101077178). Martin Kure\v{c}ka was supported by the Brno
Ph.D. Talent program funded by the Brno City Municipality.

\balance

\bibliographystyle{ACM-Reference-Format} 
\bibliography{main}

@STRING{ieee = {IEEE}}

@STRING{elsevier = {Elsevier}}

@STRING{ai = {Artificial Intelligence}}

@STRING{aaai = {{Proceedings of the AAAI Conference on Artificial Intelligence (AAAI)}}}

@STRING{aamas = {{International Conference on Autonomous Agents and Multiagent Systems (AAMAS)}}}

@STRING{cav = {{Computer Aided Verification (CAV)}}}

@STRING{iclr = {{International Conference on Learning Representations (ICLR)}}}

@STRING{icml = {{International Conference on Machine Learning (ICML)}}}

@STRING{ijcai = {{International Joint Conference on Artificial Intelligence (IJCAI)}}}

@STRING{rss = {{Robotics: Science and Systems (RSS)}}}

@STRING{tacas = {{Tools and Algorithms for the Construction and Analysis of Systems (TACAS)}}}

@STRING{uai = {{UAI}}}

@STRING{nips = {{Advances in Neural Information Processing Systems (NIPS)}}}

@inproceedings{DBLP:conf/aaai/Cubuktepe0JMST21,
  author       = {Murat Cubuktepe and
                  Nils Jansen and
                  Sebastian Junges and
                  Ahmadreza Marandi and
                  Marnix Suilen and
                  Ufuk Topcu},
  title        = {Robust Finite-State Controllers for Uncertain {POMDP}s},
  booktitle    = {{AAAI}},
  year         = {2021}
}

@inproceedings{kumar2015history,
  title={History-based controller design and optimization for partially observable {MDP}s},
  author={Kumar, Akshat and Zilberstein, Shlomo},
  booktitle={ICAPS},
  year={2015}
}

@inproceedings{DBLP:conf/ijcai/BonetG09,
  author       = {Blai Bonet and
                  Hector Geffner},
  title        = {Solving POMDPs: RTDP-Bel vs. Point-based Algorithms},
  booktitle    = {{IJCAI}},
  year         = {2009}
}

@inproceedings{pip-rnn,
  author       = {Maris F. L. Galesloot and
                  Marnix Suilen and
                  Thiago D. Sim{\~{a}}o and
                  Steven Carr and
                  Matthijs T. J. Spaan and
                  Ufuk Topcu and
                  Nils Jansen},
  title        = {Pessimistic Iterative Planning with {RNN}s for Robust {POMDP}s},
  booktitle    = {{ECAI}},
  year         = {2025},
}

@Article{kaelbling1998planning,
	author    = {Kaelbling, Leslie Pack and Littman, Michael L. and Cassandra, Anthony R.},
	title     = {Planning and acting in partially observable stochastic domains},
	journal   = ai,
	year      = {1998},
	publisher = {Elsevier}
}

@inproceedings{poupart2004bounded,
  author    = {Pascal Poupart and
Craig Boutilier},
title     = {Bounded Finite State Controllers},
booktitle = nips,
pages     = {823--830},
publisher = {{MIT} Press},
year      = {2003}
}

@article{amato2010optimizing,
	author       = {Christopher Amato and
                  Daniel S. Bernstein and
                  Shlomo Zilberstein},
  title        = {Optimizing fixed-size stochastic controllers for POMDPs and decentralized
                  POMDPs},
  journal      = {Auton. Agents Multi Agent Syst.},
  year         = {2010}
}

@inproceedings{hausknecht2015deep,
  author    = {Matthew J. Hausknecht and
Peter Stone},
title     = {Deep Recurrent {Q}-Learning for Partially Observable {MDP}s},
booktitle = {{AAAI}},
year      = {2015}
}

@inproceedings{wierstra2007solving,
	title={Solving deep memory {{POMDP}}s with recurrent policy gradients},
	author={Wierstra, Daan and F{\"o}rster, Alexander and Peters, Jan and Schmidhuber, J{\"u}rgen},
	booktitle={ICANN},
	year={2007},
}

@inproceedings{kurniawati2008sarsop,
	title={{SARSOP}: Efficient Point-Based {{POMDP}} Planning by Approximating Optimally Reachable Belief Spaces},
	author={Kurniawati, Hanna and Hsu, David and Lee, Wee Sun},
	booktitle=rss,
	publisher = {{MIT} Press},
	year      = {2008}
}

@inproceedings{junges2018finite,
	title={Finite-state controllers of {{POMDP}}s via parameter synthesis},
	author={Junges, Sebastian and Jansen, Nils and Wimmer, Ralf and Quatmann, Tim and Winterer, Leonore and Katoen, Joost-Pieter and Becker, Bernd},
	booktitle=uai,
	year      = {2018}
}

@inproceedings{carr2019counterexample,
  author    = {Steven Carr and
               Nils Jansen and
               Ralf Wimmer and
               Alexandru Constantin Serban and
               Bernd Becker and
               Ufuk Topcu},
  title     = {Counterexample-Guided Strategy Improvement for {POMDP}s Using Recurrent
               Neural Networks},
  booktitle = {{IJCAI}},
  year      = {2019},
}

@incollection{Spaan2012,
    author    = {Spaan, Matthijs T J},
    title     = {{Partially Observable {M}arkov Decision Processes}},
    booktitle = {Reinforcement Learning: State-of-the-Art},
    year      = {2012}
}

@inproceedings{DBLP:conf/ijcai/CarrJT20,
	author    = {Steven Carr and
	Nils Jansen and
	Ufuk Topcu},
	title     = {Verifiable RNN-Based Policies for {POMDP}s Under Temporal Logic Constraints},
	booktitle = {{IJCAI}},
	year      = {2020},
}

@inproceedings{DBLP:conf/ijcai/Suilen0CT20,
  author       = {Marnix Suilen and
                  Nils Jansen and
                  Murat Cubuktepe and
                  Ufuk Topcu},
  title        = {Robust Policy Synthesis for Uncertain {POMDP}s via Convex Optimization},
  booktitle    = {{IJCAI}},
  year         = {2020}
}

@inproceedings{DBLP:conf/icml/Osogami15,
  author       = {Takayuki Osogami},
  title        = {Robust partially observable {M}arkov decision process},
  booktitle    = {{ICML}},
  series       = {{JMLR} Workshop and Conference Proceedings},
  volume       = {37},
  pages        = {106--115},
  publisher    = {JMLR.org},
  year         = {2015}
}

@inproceedings{DBLP:conf/iclr/KoulFG19,
  author       = {Anurag Koul and
                  Alan Fern and
                  Sam Greydanus},
  title        = {Learning Finite State Representations of Recurrent Policy Networks},
  booktitle    = {{ICLR}},
  year         = {2019}
}

@inproceedings{DBLP:conf/icml/NiES22,
  author       = {Tianwei Ni and
                  Benjamin Eysenbach and
                  Ruslan Salakhutdinov},
  title        = {Recurrent Model-Free {RL} Can Be a Strong Baseline for Many {POMDP}s},
  booktitle    = {{ICML}},
  year         = {2022}
}

@article{DBLP:journals/neco/ZengGS93,
  author       = {Zheng Zeng and
                  Rodney M. Goodman and
                  Padhraic Smyth},
  title        = {Learning Finite State Machines With Self-Clustering Recurrent Networks},
  journal      = {Neural Computing},
  year         = {1993}
}

@article{DBLP:journals/nn/OmlinG96,
  author       = {Christian W. Omlin and
                  C. Lee Giles},
  title        = {Extraction of rules from discrete-time recurrent neural networks},
  journal      = {Neural Networks},
  volume       = {9},
  number       = {1},
  pages        = {41--52},
  year         = {1996}
}

@article{DBLP:journals/tmlr/LambrechtsBE22,
  author       = {Gaspard Lambrechts and
                  Adrien Bolland and
                  Damien Ernst},
  title        = {Recurrent networks, hidden states and beliefs in partially observable
                  environments},
  journal      = {Trans. Mach. Learn. Res.},
  year         = {2022}
}

@inproceedings{bovy2024imprecise,
  title     = {Imprecise Probabilities Meet Partial Observability: Game Semantics for Robust {POMDP}s},
  author    = {Bovy, Eline M. and Suilen, Marnix and Junges, Sebastian and Jansen, Nils},
  booktitle = {{IJCAI}},
  year      = {2024}
}

@inproceedings{DBLP:conf/nips/Hansen97,
  author       = {Eric A. Hansen},
  title        = {An Improved Policy Iteration Algorithm for Partially Observable {MDP}s},
  booktitle    = {{NIPS}},
  year         = {1997}
}

@inproceedings{DBLP:conf/ijcai/HorakBC18,
  author       = {Karel Hor{\'{a}}k and
                  Branislav Bosansk{\'{y}} and
                  Krishnendu Chatterjee},
  title        = {Goal-HSVI: Heuristic Search Value Iteration for Goal {POMDP}s},
  booktitle    = {{IJCAI}},
  year         = {2018}
}

@inproceedings{uai24,
  title={Sound Heuristic Search Value Iteration for Undiscounted POMDPs with Reachability Objectives},
  author={Ho, Qi Heng and Feather, Martin and Rossi, Federico and Sunberg, Zachary and Lahijanian, Morteza},
  booktitle={UAI},
  year={2024},
  organization={PMLR}
}

@article{tomavsek2024iterative,
  title={Iterative algorithms for solving one-sided partially observable stochastic shortest path games},
  author={Tom{\'a}{\v{s}}ek, Petr and Horak, Karel and Bo{\v{s}}ansk{\`y}, Branislav},
  journal={International Journal of Approximate Reasoning},
  year={2024},
}

@inproceedings{you2025partially,
  title={Partially observable Monte-Carlo graph search},
  author={You, Yang and Thomas, Vincent and Schutz, Alex and Skilton, Robert and Hawes, Nick and Buffet, Olivier},
  booktitle={ICAPS},
  year={2025}
}

@inproceedings{smith2004heuristic,
  title={Heuristic search value iteration for POMDPs},
  author={Smith, Trey and Simmons, Reid},
  booktitle={{UAI}},
  year={2004}
}

@inproceedings{DBLP:conf/icml/GadotWKLM24,
  author       = {Uri Gadot and
                  Kaixin Wang and
                  Navdeep Kumar and
                  Kfir Yehuda Levy and
                  Shie Mannor},
  title        = {Bring Your Own (Non-Robust) Algorithm to Solve Robust {MDP}s by Estimating
                  The Worst Kernel},
  booktitle    = {{ICML}},
  year         = {2024}
}

@article{DBLP:journals/pieee/Werbos90,
  author       = {Paul J. Werbos},
  title        = {Backpropagation through time: what it does and how to do it},
  journal      = {Proc. {IEEE}},
  year         = {1990}
}

@book{DBLP:books/wi/Puterman94,
  author       = {Martin L. Puterman},
  title        = {{Markov} Decision Processes: Discrete Stochastic Dynamic Programming},
  publisher    = {Wiley Series in Probability and Statistics},
  year         = {1994}
}

@inproceedings{DBLP:conf/cav/AndriushchenkoC21,
  author       = {Roman Andriushchenko and
                  Milan Ceska and
                  Sebastian Junges and
                  Joost{-}Pieter Katoen and
                  Simon Stupinsk{\'{y}}},
  title        = {{PAYNT:} {A} Tool for Inductive Synthesis of Probabilistic Programs},
  booktitle    = {{CAV}},
  year         = {2021}
}

@article{DBLP:journals/jair/Carr0T21,
  author       = {Steven Carr and
                  Nils Jansen and
                  Ufuk Topcu},
  title        = {Task-Aware Verifiable {RNN}-Based Policies for Partially Observable
                  {Markov} Decision Processes},
  journal      = {J. Artif. Intell. Res.},
  year         = {2021}
}

@inproceedings{hmpomdps,
  author       = {Maris F. L. Galesloot and
                  Roman Andriushchenko and
                  Milan \v{C}e\v{s}ka and
                  Sebastian Junges and
                  Nils Jansen},
  title        = {Robust Finite-Memory Policy Gradients for Hidden-Model {POMDP}s},
  booktitle    = {{IJCAI}},
  year         = {2025}
}

@inproceedings{DBLP:conf/tacas/CeskaJJK19,
  author       = {Milan Ceska and
                  Nils Jansen and
                  Sebastian Junges and
                  Joost{-}Pieter Katoen},
  title        = {Shepherding Hordes of Markov Chains},
  booktitle    = {{TACAS}},
  year         = {2019}
}

@article{madani2000undecidability,
author = {Madani, Omid and Hanks, Steve and Condon, Anne},
year = {2000},
title = {On the Undecidability of Probabilistic Planning and Infinite-Horizon Partially Observable {Markov} Decision Problems},
journal = {Proceedings of the National Conference on Artificial Intelligence}
}

@book{SuttonB:rlbook,
  title={Reinforcement Learning: An Introduction},
  author={Sutton, Richard S and Barto, Andrew G},
  year={2018},
  @@publisher={MIT Press}
}

@Article{ChatterjeeCGK16,
	author  = {Krishnendu Chatterjee and Martin Chmel{\'\i}k and Raghav Gupta and Ayush Kanodia},
	title   = {Optimal cost almost-sure reachability in {{POMDP}s}},
	journal = {Artificial Intelligence},
	year    = {2016},
}

@inproceedings{rover,
  author       = {John L. Bresina and
                  Ari K. J{\'{o}}nsson and
                  Paul H. Morris and
                  Kanna Rajan},
  @editor       = {Susanne Biundo and
                  Karen L. Myers and
                  Kanna Rajan},
  title        = {Activity Planning for the Mars Exploration Rovers},
  booktitle    = {ICAPS},
  pages        = {40--49},
  publisher    = {{AAAI}},
  year         = {2005},
}

@inproceedings{DBLP:conf/aaai/MadaniHC99,
  author    = {Omid Madani and
               Steve Hanks and
               Anne Condon},
  title     = {On the Undecidability of Probabilistic Planning and Infinite-Horizon
               Partially Observable {M}arkov Decision Problems},
  booktitle = {{AAAI/IAAI}},
  year      = {1999}
}

@article{smallwood1973optimal,
  title={The optimal control of partially observable {M}arkov processes over a finite horizon},
  author={Smallwood, Richard D and Sondik, Edward J},
  journal={Oper. Res.},
  year={1973},
  publisher={INFORMS}
}

@inproceedings{DBLP:conf/aaai/AmatoBZ10,
  author    = {Christopher Amato and
               Blai Bonet and
               Shlomo Zilberstein},
  title     = {Finite-State Controllers Based on {M}ealy Machines for Centralized and
               Decentralized {{POMDP}s}},
  booktitle = {{AAAI}},
  year      = {2010}
}

@article{STORM,
  author       = {Christian Hensel and
                  Sebastian Junges and
                  Joost{-}Pieter Katoen and
                  Tim Quatmann and
                  Matthias Volk},
  title        = {The probabilistic model checker {Storm}},
  journal      = {Int. J. Softw. Tools Technol. Transf.},
  year         = {2022}
}

@inproceedings{KNP11,
  title={{PRISM} 4.0: Verification of probabilistic real-time systems},
  author={Kwiatkowska, Marta and Norman, Gethin and Parker, David},
  booktitle={CAV},
  year={2011},
}

@InProceedings{andriushchenko23search,
author       = {Roman Andriushchenko and
                  Alexander Bork and
                  Milan Ceska and
                  Sebastian Junges and
                  Joost{-}Pieter Katoen and
                  Filip Mac{\'{a}}k},
  title        = {Search and Explore: Symbiotic Policy Synthesis in POMDPs},
  booktitle    = {{CAV}},
  year         = {2023}
}

@inproceedings{maml,
author = {Collins, Liam and Mokhtari, Aryan and Shakkottai, Sanjay},
title = {Task-robust model-agnostic meta-learning},
year = {2020},
booktitle = {NeurIPS},
}

@inproceedings{rarl,
author = {Pinto, Lerrel and Davidson, James and Sukthankar, Rahul and Gupta, Abhinav},
title = {Robust adversarial reinforcement learning},
year = {2017},
booktitle = {ICML},
}

@inproceedings{
massiani2024viability,
title={Viability of Future Actions: Robust Reinforcement Learning via Entropy Regularization},
author={Pierre-Fran{\c{c}}ois Massiani and Alexander von Rohr and Lukas Haverbeck and Sebastian Trimpe},
booktitle={EWRL},
year={2024},
}

@article{ppo_schulmann,
  author       = {John Schulman and
                  Filip Wolski and
                  Prafulla Dhariwal and
                  Alec Radford and
                  Oleg Klimov},
  title        = {Proximal Policy Optimization Algorithms},
  journal      = {CoRR},
  volume       = {abs/1707.06347},
  year         = {2017},
}

@inproceedings{
jang2017categorical,
title={Categorical Reparameterization with Gumbel-Softmax},
author={Eric Jang and Shixiang Gu and Ben Poole},
booktitle={ICLR},
year={2017},
}

@inproceedings{DBLP:journals/corr/ChoMBB14,
  title={On the Properties of Neural Machine Translation: Encoder--Decoder Approaches},
  author={Cho, Kyunghyun and van Merri{\"e}nboer, Bart and Bahdanau, Dzmitry and Bengio, Yoshua},
  booktitle={SSST},
  year={2014}
}

@Article{Musardin2022,
  author    = {Muškardin, Edi and Aichernig, Bernhard K. and Pill, Ingo and Pferscher, Andrea and Tappler, Martin},
  title     = {AALpy: an active automata learning library},
  journal   = {Innovations in Systems and Software Engineering},
  year      = {2022}}

@article{DBLP:journals/sosym/MuskardinTAP24,
  author       = {Edi Muskardin and
                  Martin Tappler and
                  Bernhard K. Aichernig and
                  Ingo Pill},
  title        = {Active model learning of stochastic reactive systems (extended version)},
  journal      = {Softw. Syst. Model.},
  year         = {2024},
}

@article{DBLP:journals/ml/MaoCJNLN16,
  author       = {Hua Mao and
                  Yingke Chen and
                  Manfred Jaeger and
                  Thomas D. Nielsen and
                  Kim G. Larsen and
                  Brian Nielsen},
  title        = {Learning deterministic probabilistic automata from a model checking
                  perspective},
  journal      = {Mach. Learn.},
  year         = {2016},

}

@inproceedings{DBLP:conf/icgi/CarrascoO94,
  author       = {Rafael C. Carrasco and
                  Jos{\'{e}} Oncina},
  editor       = {Rafael C. Carrasco and
                  Jos{\'{e}} Oncina},
  title        = {Learning Stochastic Regular Grammars by Means of a State Merging Method},
  booktitle    = {ICGI},
  year         = {1994},
}

@inproceedings{DBLP:conf/icml/WeissGY18,
  author       = {Gail Weiss and
                  Yoav Goldberg and
                  Eran Yahav},
  title        = {Extracting Automata from Recurrent Neural Networks Using Queries and
                  Counterexamples},
  booktitle    = {{ICML}},
  year         = {2018},
}

@inproceedings{DBLP:conf/cdmake/MayrY18,
  author       = {Franz Mayr and
                  Sergio Yovine},
  title        = {Regular Inference on Artificial Neural Networks},
  booktitle    = {{CD-MAKE}},
  year         = {2018},
}

@inproceedings{DBLP:conf/ifm/MuskardinAPT22,
  author       = {Edi Muskardin and
                  Bernhard K. Aichernig and
                  Ingo Pill and
                  Martin Tappler},
  title        = {Learning Finite State Models from Recurrent Neural Networks},
  booktitle    = {{IFM}},
  year         = {2022},
}

@article{DBLP:journals/sosym/AichernigKMPT24,
  author       = {Bernhard K. Aichernig and
                  Sandra K{\"{o}}nig and
                  Cristinel Mateis and
                  Andrea Pferscher and
                  Martin Tappler},
  title        = {Learning minimal automata with recurrent neural networks},
  journal      = {Softw. Syst. Model.},
  year         = {2024},
}

@Book{Kochenderfer2022,
author = {Mykel J. Kochenderfer and Tim A. Wheeler and Kyle H. Wray},
publisher = {MIT Press},
title = {Algorithms for Decision Making},
year = {2022},
}

@inproceedings{rlhf_2022,
author = {Ouyang, Long and Wu, Jeff and Jiang, Xu and Almeida, Diogo and Wainwright, Carroll L. and Mishkin, Pamela and Zhang, Chong and Agarwal, Sandhini and Slama, Katarina and Ray, Alex and Schulman, John and Hilton, Jacob and Kelton, Fraser and Miller, Luke and Simens, Maddie and Askell, Amanda and Welinder, Peter and Christiano, Paul and Leike, Jan and Lowe, Ryan},
title = {Training language models to follow instructions with human feedback},
year = {2022},
booktitle = {NeurIPS},
}

@article{lstm_1997, title={Long Short-Term Memory}, volume={9}, ISSN={0899-7667},number={8}, journal={Neural Computation}, author={Hochreiter, Sepp and Schmidhuber, Jürgen}, year={1997}, month=nov, pages={1735–1780} }

@software{jax2018github,
  author = {James Bradbury and Roy Frostig and Peter Hawkins and Matthew James Johnson and Chris Leary and Dougal Maclaurin and George Necula and Adam Paszke and Jake Vander{P}las and Skye Wanderman-{M}ilne and Qiao Zhang},
  title = {{JAX}: composable transformations of {P}ython+{N}um{P}y programs},
  url = {http://github.com/jax-ml/jax},
  version = {0.3.13},
  year = {2018},
}

@misc{TFAgents,
  title = {{TF-Agents}: A library for Reinforcement Learning in TensorFlow},
  author = {Sergio Guadarrama and Anoop Korattikara and Oscar Ramirez and
     Pablo Castro and Ethan Holly and Sam Fishman and Ke Wang and
     Ekaterina Gonina and Neal Wu and Efi Kokiopoulou and Luciano Sbaiz and
     Jamie Smith and Gábor Bartók and Jesse Berent and Chris Harris and
     Vincent Vanhoucke and Eugene Brevdo},
  howpublished = {\url{https://github.com/tensorflow/agents}},
  url = "https://github.com/tensorflow/agents",
  year = 2018,
  note = "[Online; accessed 25-June-2025]"
}

@inproceedings{gleave_2021,
  title={Adversarial Policies: Attacking Deep Reinforcement Learning},
  author={Gleave, Adam and Dennis, Michael and Wild, Cody and Kant, Neel and Levine, Sergey and Russell, Stuart},
  booktitle={ICLR},
  year={2021},
 

}

@article{huang_2017, title={Adversarial Attacks on Neural Network Policies}, author={Huang, Sandy and Papernot, Nicolas and Goodfellow, Ian and Duan, Yan and Abbeel, Pieter}, year={2017}, journal="CoRR"}

@inproceedings{DBLP:conf/iclr/ZhangCBH21,
  author       = {Huan Zhang and
                  Hongge Chen and
                  Duane S. Boning and
                  Cho{-}Jui Hsieh},
  title        = {Robust Reinforcement Learning on State Observations with Learned Optimal
                  Adversary},
  booktitle    = {{ICLR}},
  year         = {2021}
}

@inproceedings{liu_2023,
  title={On the Robustness of Safe Reinforcement Learning under Observational Perturbations},
  author={Liu, Zuxin and Guo, Zijian},
  year={2023},
  booktitle={{ICLR}}
}

@INPROCEEDINGS{5683311,
  author={Fei, Xin and Boukerche, Azzedine and Yu, Richard},
  booktitle={GLOBECOM}, 
  title={A POMDP Based K-Coverage Dynamic Scheduling Protocol for Wireless Sensor Networks}, 
  year={2010},
}

@article{annurev:/content/journals/10.1146/annurev-control-042920-092451,
   author = "Kurniawati, Hanna",
   title = "Partially Observable Markov Decision Processes and Robotics", 
   journal= "Annual Review of Control, Robotics, and Autonomous Systems",
   year = "2022",
  }

@Article{healthcare10020283,
AUTHOR = {Zhang, Wenqian and Wang, Haiyan},
TITLE = {Diagnostic Policies Optimization for Chronic Diseases Based on POMDP Model},
JOURNAL = {Healthcare},
YEAR = {2022},
}

@article{DBLP:journals/nature/WurmanBKMS0CDE022,
  author       = {Peter R. Wurman and
                  Samuel Barrett and
                  Kenta Kawamoto and
                  James MacGlashan and
                  Kaushik Subramanian and
                  Thomas J. Walsh and
                  Roberto Capobianco and
                  Alisa Devlic and
                  Franziska Eckert and
                  Florian Fuchs and
                  Leilani Gilpin and
                  Piyush Khandelwal and
                  Varun Raj Kompella and
                  HaoChih Lin and
                  Patrick MacAlpine and
                  Declan Oller and
                  Takuma Seno and
                  Craig Sherstan and
                  Michael D. Thomure and
                  Houmehr Aghabozorgi and
                  Leon Barrett and
                  Rory Douglas and
                  Dion Whitehead and
                  Peter D{\"{u}}rr and
                  Peter Stone and
                  Michael Spranger and
                  Hiroaki Kitano},
  title        = {Outracing champion Gran Turismo drivers with deep reinforcement learning},
  journal      = {Nature},
  year         = {2022},
}

@inproceedings{
dohare2023overcoming,
title={Overcoming Policy Collapse in Deep Reinforcement Learning},
author={Shibhansh Dohare and Qingfeng Lan and A. Rupam Mahmood},
booktitle={Sixteenth European Workshop on Reinforcement Learning},
year={2023},
}

@article{loss_of_plasticity,
       author = {{Dohare}, Shibhansh and {Hernandez-Garcia}, J. Fernando and {Lan}, Qingfeng and {Rahman}, Parash and {Mahmood}, A. Rupam and {Sutton}, Richard S.},
        title = "{Loss of plasticity in deep continual learning}",
      journal = {Nature},
         year = 2024,
        month = aug,
       volume = {632},
       number = {8026},
        pages = {768-774},
}

@article{challenges_rl,
author = {Dulac-Arnold, Gabriel and Levine, Nir and Mankowitz, Daniel J. and Li, Jerry and Paduraru, Cosmin and Gowal, Sven and Hester, Todd},
title = {Challenges of real-world reinforcement learning: definitions, benchmarks and analysis},
year = {2021},
journal = {Mach. Learn.},
}

@inproceedings{10.5555/2073796.2073844,
author = {Meuleau, Nicolas and Kim, Kee-Eung and Kaelbling, Leslie Pack and Cassandra, Anthony R.},
title = {Solving POMDPs by searching the space of finite policies},
year = {1999},
booktitle = {UAI},
}

@misc{implementation,
author={David Hud\'ak and Maris F. L. Galesloot and Martin Tappler and Martin Kure\v{c}ka and Nils Jansen and Milan \v{C}e\v{s}ka},
title="Lexpop Synthesis Repository (AAMAS Camera-Ready Branch)",
url={https://github.com/DaveHudiny/rl_synthesis/tree/AAMAS-Camera-Ready},
year={2026}
}

@article{DBLP:journals/automatica/ZhengJL22,
  author       = {Wei Zheng and
                  Taeho Jung and
                  Hai Lin},
  title        = {The Stackelberg equilibrium for one-sided zero-sum partially observable
                  stochastic games},
  journal      = {Automatica},
  year         = {2022}
}

@InProceedings{pmlr-v80-verma18a,
  title = 	 {Programmatically Interpretable Reinforcement Learning},
  author =       {Verma, Abhinav and Murali, Vijayaraghavan and Singh, Rishabh and Kohli, Pushmeet and Chaudhuri, Swarat},
  booktitle = 	 {Proceedings of the 35th International Conference on Machine Learning},
  year = 	 {2018},
  publisher =    {PMLR},

}

@article{DBLP:journals/tac/WintererJWJTKB21,
  author       = {Leonore Winterer and
                  Sebastian Junges and
                  Ralf Wimmer and
                  Nils Jansen and
                  Ufuk Topcu and
                  Joost{-}Pieter Katoen and
                  Bernd Becker},
  title        = {Strategy Synthesis for POMDPs in Robot Planning via Game-Based Abstractions},
  journal      = {{IEEE} Trans. Autom. Control.},
  year         = {2021}
}

@inproceedings{mepomdpnips25,
  author = {Bovy, Eline M. and Probine, Caleb and Suilen, Marnix and Topcu, Ufuk and Jansen, Nils},
  title = {Multi-Environment {POMDP}s: Discrete Model Uncertainty Under Partial Observability},
  booktitle = {{NeurIPS}},
  year = {2025}
}

@article{DBLP:journals/sttt/WintererWBJ24,
  author       = {Leonore Winterer and
                  Ralf Wimmer and
                  Bernd Becker and
                  Nils Jansen},
  title        = {Strong Simple Policies for POMDPs},
  journal      = {Int. J. Softw. Tools Technol. Transf.},
  year         = {2024}
}

@inproceedings{rpomdpnips25,
  author = {Krale, Merlijn and Bovy, Eline M. and Galesloot, Maris F. L. and Sim{\~{a}}o, Thiago D. and Jansen, Nils},
  title = {On Evaluating Policies for Robust {POMDP}s},
  booktitle = {{NeurIPS}},
  year = {2025},
}

\newpage

\appendix
\onecolumn
   \section{Benchmark Models}\label{appendix:model_setting}
   
As we mention in the experiment~\Cref{sec:exp}, we took the single-POMDP models from the \saynt repository~\cite{andriushchenko23search, DBLP:conf/cav/AndriushchenkoC21}, and the models for the robust setting from the \rfpg paper~\cite{hmpomdps}, where most details of those models are discussed.

In the single-POMDP setting, we scaled the \texttt{intercept} and \texttt{evade} models by increasing the parameters governing model size, thereby significantly increasing the number of states. The remaining models are identical to those introduced in the \saynt repository~\cite{andriushchenko23search}.

To extend our set of HM-POMDP benchmarks, we scaled the \texttt{avoid} model using the available variables to determine the model size and the possible locations of the oscillating obstacles. The drone model is extended by a) adding a slippery effect to the agent's behavior and b) adding more potential initial positions of the drones. The case of the moving obstacle model is similar -- we enlarged the area by adding new holes where obstacles can spawn and added a small slippery probability to each static obstacle, which significantly increases the state space and makes the model more challenging.

For reproducibility, all models used are included in the source code~\cite{implementation}.

   \section{DRL Details}\label{appendix:rl_section}
   
\subsection{Reward Shaping}
Since all the models come with a reachability specification (from the PRISM language) that does not correspond to the RL reward-oriented setting, we had to implement our own reward function to allow the agents to optimize over given models. 
We perform \emph{reward shaping} to optimize the DRL models.
Importantly, \textbf{reward shaping is not used in the evaluation}; there, whether evaluated empirically or on the full model, we measure the value defined by the PRISM model.
Thus, all reported values come from the same reward model.

In the \emph{reachability} setting, a sparse reward of $40$ is given once the agent reaches the target. In contrast, in cases where reaching the goal state is easy and the main objective is {reachability reward maximization}, we give only a small reward for reaching the goal state ($2$ in case of minimizing, $1$ in case of maximizing), and the reward from the environment is multiplied by $(-)10.0$. We do not use negative rewards for failing the task, as it usually leads to too conservative policies.\Cref{tab:simple_reward_models} summarizes which model belongs to which category in the case of simple reachability and reward models.

\begin{table}[!htp]
    \centering
    \begin{tabular}{cc}
    \toprule
         Model & Reward Type \\
    \midrule
         maze-10 & Reward Maximizing \\
         rocks-16 & Reward Minimizing \\
         network-3-8-20 & Reward Minimizing \\
         network-5-10-8 & Reward Minimizing \\
         intercept-16 & Reachability Maximizing \\
         evade-n17 & Reachability Maximizing \\
         drone-2-8-1 & Reachability Maximizing \\
    \midrule
         network & Reward Maximizing \\
         drone-2-6-1 & Reachability Maximizing \\
    \bottomrule
    \end{tabular}
    \caption{Taxonomy of our reachability and rewards models. The top part contains the single-POMDP models, while the bottom part describes the models from the robust setting. Models with more specific reward functions are described in~\Cref{tab:combined_reward_models}.}
    \label{tab:simple_reward_models}
\end{table}

However, in the case of the HM-POMDP setting, the reward is usually more complicated, as we cannot guarantee reaching the goal state and still want to optimize the reward. 
In this case, the DRL performance strongly depends on the designed reward: when the reward is too small, the agent may do nothing, and when it is too large, the agent may ignore important aspects of the model. For these models, we use a combination of a multiplier for rewards from the environment, a reward for reaching the goal state, and a negative penalty for truncating the episode. We describe these rewards in~\Cref{tab:combined_reward_models}.

\begin{table}
    \centering
    \begin{tabular}{cccc}
    \toprule
         Model & Goal Reward & Reward Multiplier & Truncation Penalty \\
    \midrule
         avoid & 400 & -1.0 & -10.0 \\
         (moving-)obstacles & 360 & -1.0 &  0.0 \\
         rover & 160 & 1.0 & 0.0 \\
    \bottomrule
    \end{tabular}
    \caption{Specific reward models for specific HM POMDPs.}
    \label{tab:combined_reward_models}
\end{table}

Furthermore, to motivate the agent to explore more, we use the configurable entropy regularization, which is useful when the reward is sparse and the agent would, by default, converge to some local optima. This approach is useful in terms of simple general implementation without configuring the reward for each model independently.

\subsection{Vectorized Simulation}\label{appendix:vecstorm}
PPO is an on-policy algorithm and, as such, requires a large number of on-policy samples to ensure stability and fast convergence. In our experiments, the use of \storm as a simulator became a runtime bottleneck, since \storm is primarily designed for model verification. Although it supports sampling from the simulator, running multiple environment instances in parallel was too resource-intensive. To address this, we implemented a lightweight simulator for \storm environments that supports vectorized execution of multiple parallel environments. The simulator uses \storm to parse the environment specification file and generate sparse tensors that describe the environment dynamics. These tensors are sufficient to simulate the environment efficiently. Additionally, we employ the JAX just-in-time~(JIT) compilation utility to compile the generic transition function for the specific dynamics tensor of each environment, thereby further improving the simulation speed.

\subsection{RL Library}\label{appendix:library}
To implement reinforcement learning, we used the TensorFlow Agents library, primarily because it supports recurrent neural network policies (unlike the popular choice Stable Baselines~3\footnote{\href{https://stable-baselines3.readthedocs.io/en/master/}{https://stable-baselines3.readthedocs.io/en/master/}}) and achieved the best performance among the libraries we tested in preliminary experiments. 

The framework provides multiple recent deep reinforcement learning algorithms, but in our case, we use only the PPO algorithm with an LSTM recurrent neural network because our preliminary results showed the best and most stable overall performance from the set of implemented algorithms (DQN, DDQN, PPO, and discretized SAC). We should note that the main concern of the paper is not the selection of the best learning algorithm; PPO is a relatively standardized and generally available algorithm implemented in most RL libraries. 
In general, our framework can work with any RL algorithm for POMDPs.

\subsection{Reinforcement Learning Network Architectures}\label{appendix:rl_architecture}
For reinforcement learning, we use the standard LSTM actor-critic architecture introduced in the TensorFlow Agents framework. The architecture contains a single dense input; the network consists of an LSTM encoder with always the same architecture, followed by a task-dependent dense categorical projection network in the case of the actor or by a single neuron representing a value layer in the case of the critic. The fixed architecture of the LSTM encoder consists of three layers -- one dense input layer with 64 neurons, one LSTM layer with size 32, and one output layer with size 64. We experimented with larger neural networks, but in general, the size of the network was sufficient for all benchmarks.

\subsection{Self-Interpretation with Deep RL}
\label{app:sig_drl}
In our preliminary experiments, we tried to combine PPO training with the SIG actor. However, it was shown that the discrete structure significantly affects exploration, and the convergence is much slower than with the RNN-based PPO. Thus, we chose the more effective approach, where we allow exploration of POMDPs using RNN-based PPO and then estimate the policy using SIG extraction.

\subsection{Dynamic Action Space}\label{appendix:dynamic_action_space}
One important aspect of the benchmark models in this paper is that they usually have a dynamic action space, which allows only a subset of actions to be played in each state. 
This is problematic from the perspective of deep reinforcement learning, where we typically have a fixed number of output nodes that determine the categorical distribution over actions for each observation-history tuple. 
We found that if we use a wrapper over the PPO algorithm that completely removes the possibility of playing illegal actions, the learning sometimes completely breaks down when we sample from a distribution different from the one the agent produces, thereby violating the on-policy setting of the PPO algorithm. 
We experimented with multiple options, but the best setting we found is to allow all actions to be played and to generate random actions played in the simulator without letting the agent know. This setting partially limits the agent.

We also explored the options of stochastic and deterministic policies. Our preliminary experiments show that both stochastic exploration and stochastic evaluation achieve the best results in both single-POMDP and hidden-model POMDP settings, with the random replacement of illegal actions on the simulator side. However, for extraction, we sampled from a masked policy, i.e., we removed illegal actions, which slightly affected performance but also made extraction more stable. However, more research might be beneficial in this area.

\subsection{Robust \toolname Details}\label{appendix:robust_learning}
The robust training is ensured by our vectorized simulator. As we show in an~\Cref{alg:lexpop}, we initialize the training with a first POMDP\footnote{Technically, we use more POMDPs for the first iterations to make the worst-case selection loop faster.}, which we add to the simulator and then perform a standard reinforcement learning loop with our highly vectorized simulator. In the first iteration, all 256 subsimulators simulate the initial POMDP. When we add a new POMDP, we recompute the resources for a new simulator using geometric progression. The number of simulators for the last POMDP is
\begin{equation*}
    a_L = \frac{L (1-q)}{1-q^S}
\end{equation*}
simulators and each following POMDP has $a_i = q a_{i+1}$ simulators from $S$ available simulators, where the geometric coefficient $q$ has a constant value $0.4$ and i: $1 \leq i \leq L-1$ is an index of simulator added in iteration $i$ with the current last iteration $L$. This mechanism resembles a momentum, where we mostly optimize for the most recent POMDP while still remembering parts of the gradient from the past. 

To start training in the first iteration, we initialize a single agent that we later transfer across all training iterations, including both the actor and the critic components. The extraction in the case of Alergia does not share any information between independent outer iterations of the Robust \toolname (Line 12 of the~\Cref{alg:lexpop}), whereas the extraction using the SIG method shares the original self-interpretable network among multiple iterations to achieve more stable performance with faster extractions, which was clearly beneficial in the moving-obstacles task, as shown in~\Cref{fig:curves}. 

To decrease the impact of loss of plasticity phenomena mentioned in~\cite{dohare2023overcoming, loss_of_plasticity}, we perform inner training with only a limited number of iterations, use the non-stationary version of Adam, and use small weight decay, as we summarize in~\Cref{tab:ppo_parameters}.

\subsection{Parameter Discussion and Summary}
One of the typical issues with reinforcement learning algorithms is the large number of configurable parameters. In our paper, we primarily used the default parameters of the TensorFlow Agents library, along with our empirical observations from preliminary experiments.

\paragraph{Number of Inner and Outer Training Iterations}
In the single-POMDP setting, we use a fixed number of 4000 iterations of the \toolname algorithm, which was sufficient to achieve stable performance across all benchmark models and multiple seeds, with most models achieving close-to-final performance before the first 1000 iterations. In the case of hidden model POMDPs, the number of DRL training iterations has a more significant effect on performance, since more inner iterations usually lead to fewer outer extraction iterations of the Robust \toolname algorithm and to potential over-optimization for specific POMDPs rather than robust behavior. While the rest of the parameters were exactly the same in all of the environments, we used three model-specific parameters: a) number of initial training iterations, which affects how strictly the model learns the initial model(s) and depends on our empirical experience of how long; b) number of inner training iterations, which affects how much the agent optimizes for new POMDPs; and c) number of initial (random) POMDPs for which the agent starts optimizing. We show the setting in~\Cref{tab:training_iters}. The number of initial POMDPs was 6 globally, whereas in the avoid-large case, we found an improvement with 11 initial POMDPs. 

\begin{table}[!htp]
    \centering
    \begin{tabular}{ccc}
    \toprule
         Model & $\#$ init. iter. & $\#$ inner iter. \\
    \midrule
         rover & 400 & 100 \\
         obstacles-8-5 & 400 & 70 \\
         network & 400 & 150 \\
         maze-10 & 400 & 150 \\
         avoid & 200 & 25 \\
         avoid-large & 250 & 35 \\
         drone-2-6-1 & 400 & 50 \\
         moving-obstacles & 400 & 50 \\
    \bottomrule
    \end{tabular}
    \caption{Training iterations for each model.}
    \label{tab:training_iters}
\end{table}

\paragraph{Discounting}
The formal specifications of our models are undiscounted, which is in contrast to how reinforcement learning works, since no discounting leads to unstable learning that breaks after several training iterations. In our preliminary experiments, we tried to set the discounting to the values of the set $\{0.99, 0.995, 0.999\}$. Although the first value significantly changed the optimal solution, the latest option was unstable, and the learning did not work correctly. In general, in all our experiments, we use a discount rate of $0.995$.

\paragraph{Summary}
In the following two tables~\Cref{tab:training_parameters} and~\Cref{tab:ppo_parameters}, we summarize all the non-default parameters of our implementation. The rest uses the default parameters. from the TensorFlow Agents framework, mostly taken from the original PPO paper~\cite{ppo_schulmann}. 

\begin{table}[!htp]
    \centering
    \begin{tabular}{cc}
    \toprule
         Parameter & Value \\
         \midrule
         Batch Size & 256  \\
         Parallel Training Simulators & 256 \\
         Parallel Evaluation Simulators & 512 \\
         Learning Trajectory Length & 32 \\
         Maximum Number of Steps & 601 \\
    \bottomrule
    \end{tabular}
    \caption{Sampling parameters of the RL Setting.}
    \label{tab:training_parameters}
\end{table}
\begin{table}[!htp]
    \centering
    \begin{tabular}{cc}
    \toprule
         Parameter & Value \\
    \midrule
         Importance Ratio Clipping & 0.2 \\
         Num PPO Epochs & 3 \\
         Advantage Estimator & GAE \\
         (GAE) Lambda & 0.95 \\
         Entropy Regularization & 0.02 \\
         Normalize Rewards & True \\
         Normalize Observations & True \\
         Discount & 0.995 \\
         Optimizer & Adam \\
         Learning Rate & 0.00016 \\
         Adam $\beta_1$ & 0.99 \\
         Adam $\beta_2$ & 0.99 \\
         Weight Decay (L2 Reg.) & 0.0001 \\ 
    \bottomrule
    \end{tabular}
    \caption{Parameters of PPO. The rest parameters use the default TensorFlow Agents values, which follow the original PPO paper~\cite{ppo_schulmann}.} 
    \label{tab:ppo_parameters}
\end{table}

\section{Extraction Details using Self-Interpretation}

\subsection{Extraction}
For extraction, we use two options as described in \cref{sec:fsc-extraction}: the automata-learning-based AALpy library using the Alergia algorithm, and machine-learning-based self-interpretable Gumbel networks (SIG). Both methods use the same amount of data, and in the single-POMDP setting, we use the same policy and the same sampled data for both extractions. The input for all methods was the same -- 4001 steps in 256 parallel environments, i.e., over a million steps in the environment usable for automata learning, Alergia, or cloning-oriented SIG. In the case of Alergia, we did not tune any configurable parameters, whereas in the case of the SIG extraction, we use a variable number of training epochs for cloning and an architectural upper bound that limits the maximum FSC size. The configuration is summarized in~\Cref{tab:extraction_parameters}.

\begin{table}[!htp]
    \centering
    \begin{tabular}{cccc}
    \toprule
         Parameter & Large Robust & Small Robust & Single \\
    \midrule
         Training Epochs &  5001 & 501 & 6001 \\
         Max FSC Size & 10.0 & 3.0 & 3.0 \\
    \bottomrule
    \end{tabular}
    \caption{Self-interpretation SIG setting. We describe the setting for the large robust models (avoid-large, drone-2-6-1, moving-obstacles), small robust models (avoid, network, obstacles-8-5, rover), and for the single-POMDP setting.}
    \label{tab:extraction_parameters}
\end{table}

\subsection{Gumbel Softmax Self-Interpretable Networks (SIG) Architecture}\label{appendix:architectures}
Since we use a custom architecture for SIG interpretation, we show its complete architecture in~\Cref{fig:sig_architecture}. If we compare it with the architecture described in~\Cref{appendix:rl_architecture}, the network has more layers and larger layer sizes than the reinforcement learning network. We use the larger architecture to ensure the network has greater capacity to encode the policy without using memory and to use recurrent feedback when necessary. The inner GRU network uses (recurrent) dropout with a value $0.2$ to not overuse memory, which is limited to only a maximum size $k$. During extraction, the network does not necessarily use all the memory it is provided, so in the case of our main robust experiment, in~\Cref{tab:robust_results}, we describe only the size of the reachable memory from an initial memory observation tuple. The upper bounds on memory for a given set of benchmarks are described in~\Cref{tab:extraction_parameters}, where we separate the large models from the HM POMDP setting, where more memory is beneficial, and the smaller HM POMDPs and single POMDP models, where we found that agents can be approximated with smaller memory. 

In the case of the avoid-large experiment with the sole GRU network, as we depict in~\Cref{fig:curves}, we removed all bottom layers and used only the top part with the same GRU layer, now giving the hidden state after each step directly to itself. 

\begin{figure}
    \centering
    \includegraphics[width=0.45\linewidth]{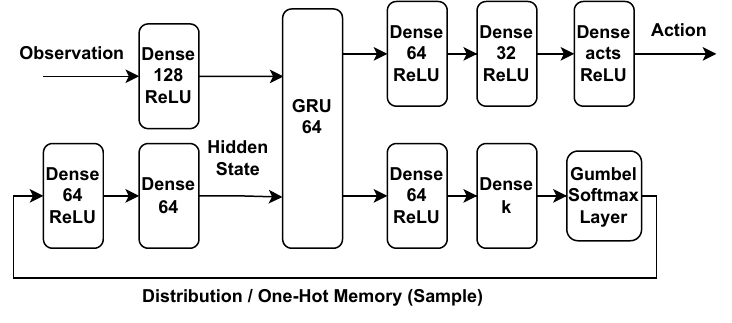}
    \caption{The architecture of our Gumbel Softmax self-interpretable network. The upper part of each block describes the architecture of a layer, the second number represents the size of the layer, where $k$ is the upper bound of memory, and $\text{acts}$ is the number of playable actions for a given model. During the training, we use the output distribution from the Gumbel layer, while in the case of the final controller construction, we use a one-hot vector representing the sampled memory.}
    \label{fig:sig_architecture}
\end{figure}

\subsection{Gumbel Softmax Layer}
To sample the one-hot vector $y$ from a categorical distribution  $\softmax(p)$ so that $y$ is differentiable with respect to $p$, a reparameterization is needed. A classical approach is to sample a vector $g \in \mathbb{R}^{\lvert p \rvert}$ of i.i.d. samples from the standard Gumbel distribution, and set $y$ as the one-hot encoding of $\arg\max_i(p_i+g_i)$. It is known that the distribution of $\arg\max_i(p_i+g_i)$ is indeed $\softmax(p)$. However, this approach has a nonzero gradient only with respect to the maximizing $p_i$, which is the reason we opted for the Gumbel Softmax reparameterization~\cite{jang2017categorical}:

\begin{equation*}
    y = \softmax((p + g)/\tau)
\end{equation*}
where $\tau$ is a temperature, and the values $g_i$ are again independent random samples from the standard Gumbel distribution. 

In the limit, as $\tau$ goes to zero, $y$ coincides with the one-hot encoding of a random sample drawn from the categorical distribution $\softmax(p)$.
On the other hand, for a fixed $\tau$, the function $y$ is continuous and differentiable with respect to the whole $p$, which improves the stability of the training. 
We use cosine scheduling to reduce the temperature~$\tau$ during training
that yields a distribution close to the original categorical distribution.
   \section{Evaluation Methodology}\label{appendix:stats_section}
   \subsection{Empirical Evaluation}\label{appendix:evaluation}
The neural network policies were evaluated on 600 steps in 512 parallel environments.
The truncated trajectories were considered as ending in a fail state.
This gives us confident results, as we usually have over two thousand evaluation episodes for evaluation. For instance, we can see that the empirical values of the extracted finite-state controllers are close to their verified values.

\subsection{Statistical Tests}

To compare the algorithms in Table~1 and Table~2, we used the Sign test and the Brunner Munzel test. The Sign test serves to compare the median of a random variable $X$ with a fixed value $m$ without assumptions on the distribution of $X$. Its null and alternative hypothesis are as follows:
\begin{itemize}
    \item[$H_0$:] $P(X > m) = 0.5$
    \item[$H_1$:] $P(X > m) > 0.5$
\end{itemize}
The test was used to compare \saynt with other methods as the performance of \saynt is deterministic.

The Brunner-Munzel test serves to compare two random variables $X$ and $Y$ under the assumptions they are independent. Its null and alternative hypothesis are as follows:
\begin{itemize}
    \item[$H_0$:] $P(X > Y) = P(Y > X)$
    \item[$H_1$:] $P(X > Y) > P(Y > X)$
\end{itemize}
The test was used to compare the results of the two extraction methods.

The significant test results are summarized in \Cref{tab:tests-1} and \Cref{tab:tests-2}.

\begin{table}[t]
\begin{tabular}{lllr}
\toprule
benchmark & $X$ & $Y$ & $p$-value \\
\midrule
rocks-16 & \saynt & LP+\sig & 9.8e-04 \\
rocks-16 & \saynt & LP+\alergia & 9.8e-04 \\
rocks-16 & LP+\sig & LP+\alergia & 1.4e-05 \\
network-3-8-20 & LP+\sig & \saynt & 9.8e-04 \\
network-3-8-20 & LP+\alergia & \saynt & 9.8e-04 \\
network-3-8-20 & LP+\sig & LP+\alergia & 4.5e-06 \\
network-5-10-8 & LP+\sig & \saynt & 9.8e-04 \\
network-5-10-8 & LP+\alergia & \saynt & 9.8e-04 \\
network-5-10-8 & LP+\sig & LP+\alergia & 1.6e-03 \\
intercept-16 & LP+\sig & \saynt & 9.8e-04 \\
intercept-16 & LP+\alergia & \saynt & 9.8e-04 \\
intercept-16 & LP+\sig & LP+\alergia & 5.9e-03 \\
evade-n17 & LP+\sig & \saynt & 9.8e-04 \\
evade-n17 & LP+\alergia & \saynt & 9.8e-04 \\
drone-2-8-1 & LP+\sig & \saynt & 9.8e-04 \\
drone-2-8-1 & LP+\alergia & \saynt & 9.8e-04 \\
\bottomrule
\end{tabular}
\caption{The statistically significant comparisons of the methods in Table~1 and their p-values. The methods LP+\sig and LP+\alergia represent \toolname with the respective extraction strategy. The tests reject $P(X>Y) = P(Y>X)$ in favor of $P(X>Y) > P(Y>X)$.}
\label{tab:tests-1}
\end{table}

\begin{table}
\begin{tabular}{lllr}
\toprule
benchmark & $X$ & $Y$ & $p$-value \\
\midrule
rover & \rfpg & LP+\sig & 3.6e-06 \\
rover & \rfpg & LP+\alergia & 7.1e-13 \\
rover & LP+\sig & LP+\alergia & 4.2e-03 \\
obstacles-8-5 & \rfpg & LP+\sig & 3.4e-12 \\
obstacles-8-5 & \rfpg & LP+\alergia & 7.1e-13 \\
network & LP+\sig & \rfpg & 1.8e-06 \\
network & LP+\alergia & \rfpg & 1.8e-06 \\
avoid & LP+\alergia & \rfpg & 4.1e-02 \\
avoid-large & LP+\sig & \rfpg & 7.8e-04 \\
avoid-large & LP+\alergia & \rfpg & 2.8e-03 \\
drone-2-6-1 & LP+\sig & \rfpg & 7.1e-13 \\
drone-2-6-1 & LP+\alergia & \rfpg & 5.5e-12 \\
moving\_obstacles & LP+\sig & \rfpg & 7.1e-13 \\
moving\_obstacles & LP+\alergia & \rfpg & 1.6e-03 \\
moving\_obstacles & LP+\sig & LP+\alergia & 7.1e-13 \\
\bottomrule
\end{tabular}
\caption{The statistically significant comparisons of the methods in Table~2 and their p-values. The methods LP+\sig and LP+\alergia represent \toolname with the respective extraction strategy. The tests reject $P(X>Y) = P(Y>X)$ in favor of $P(X>Y) > P(Y>X)$.}
\label{tab:tests-2}
\end{table}
   \section{Extended Results}\label{appendix:extended_results}
   In this section, we show all the convergence curves for both the single-POMDP and the hidden-model POMDP setting.

\subsection{Single-POMDP Experiments}
\begin{figure}
    \centering
    \begin{subfigure}{.24\textwidth}
    \centering
    \includegraphics[width=.95\linewidth]{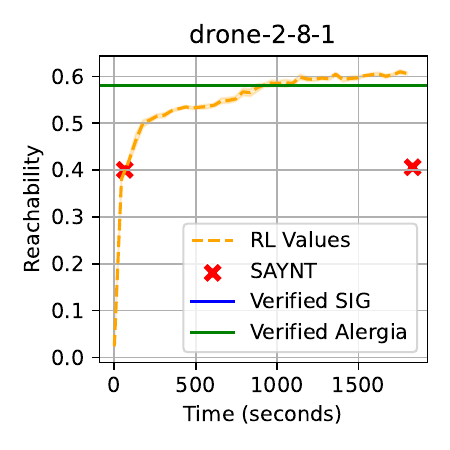}
    \caption{Drone-2-8-1}
    \end{subfigure}
    \begin{subfigure}{.24\textwidth}
    \centering
    \includegraphics[width=.95\linewidth]{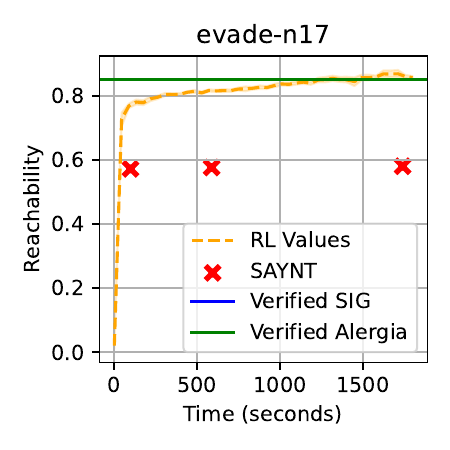}
    \caption{Evade-n17}
    \end{subfigure}
    \begin{subfigure}{.24\textwidth}
    \centering
    \includegraphics[width=.95\linewidth]{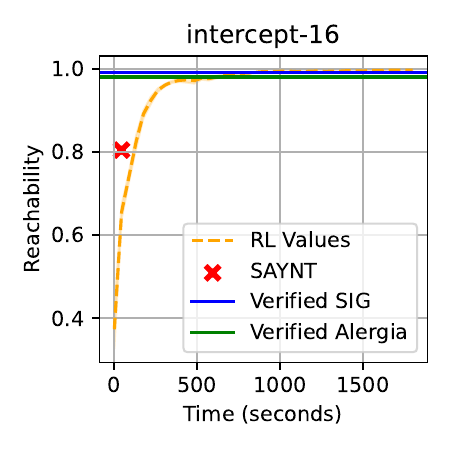}
    \caption{Intercept-16}
    \end{subfigure}
    \begin{subfigure}{.24\textwidth}
    \centering
    \includegraphics[width=.95\linewidth]{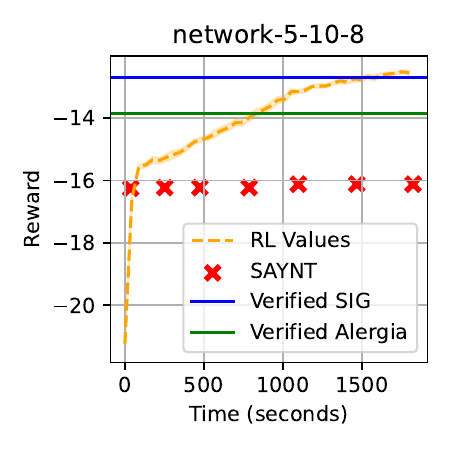}
    \caption{Network-5-10-8}
    \end{subfigure}
    \begin{subfigure}{.24\textwidth}
    \centering
    \includegraphics[width=.95\linewidth]{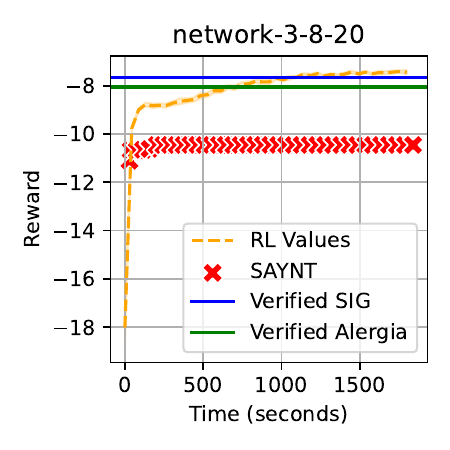}
    \caption{Network-3-8-20}
    \end{subfigure}
    \begin{subfigure}{.24\textwidth}
    \centering
    \includegraphics[width=.95\linewidth]{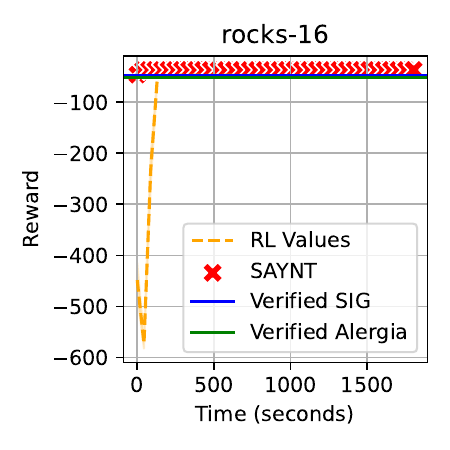}
    \caption{Rocks-16}
    \end{subfigure}
    \begin{subfigure}{.24\textwidth}
    \centering
    \includegraphics[width=.95\linewidth]{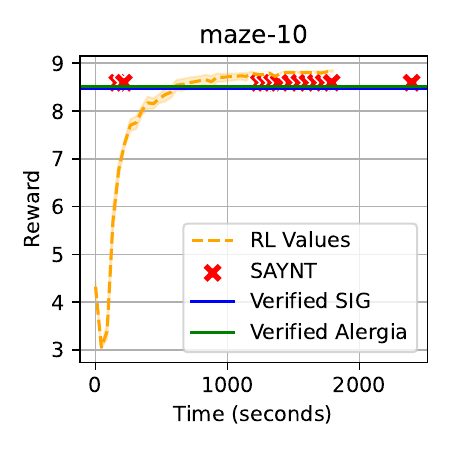}
    \caption{Maze-10}
    \end{subfigure}
    \caption{Figures for single-POMDP models. The yellow line shows the RL performance over time, red crosses the values provided by \saynt over time, the blue line the final verified value extracted by SIG, and the green line the verified value extracted by Alergia. }
    \label{fig:all_single_pomdps}
\end{figure}

\Cref{fig:all_single_pomdps} shows the training of our RL algorithm before extraction, the mean verified values of the extracted controllers by both our methods (Alergia and SIG), and the development of the value of the controllers synthesized by \saynt. 

\subsection{Hidden Model POMDP Experiments}
\begin{figure}
    \centering
    \begin{subfigure}{.24\textwidth}
    \centering
    \includegraphics[width=.95\linewidth]{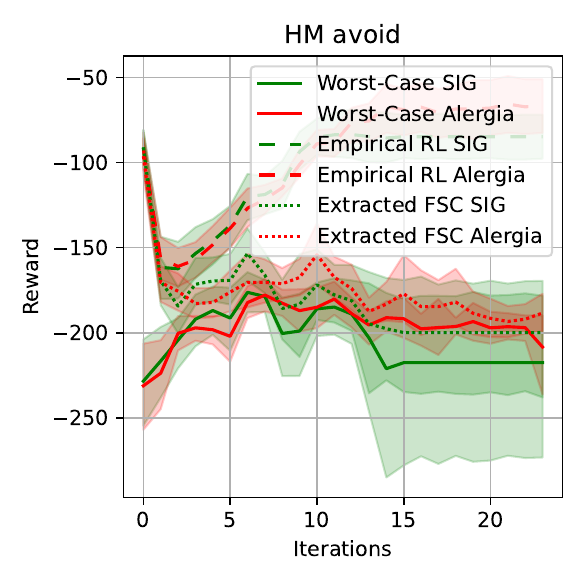}
    \caption{Avoid}
    \end{subfigure}
    \begin{subfigure}{.24\textwidth}
    \centering
    \includegraphics[width=.95\linewidth]{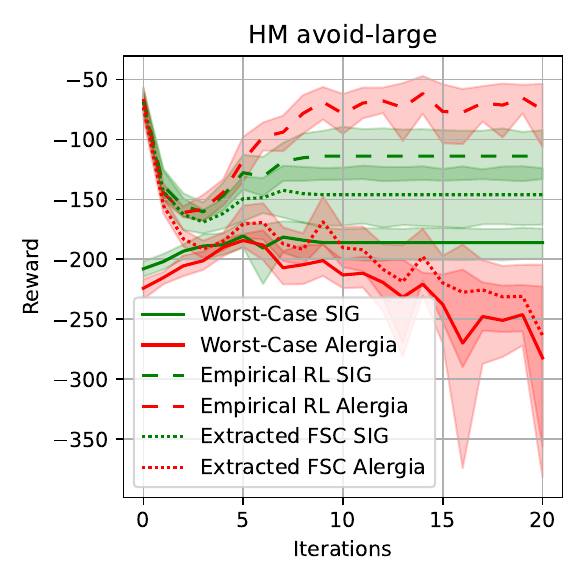}
    \caption{Avoid-Large}
    \end{subfigure}
    \begin{subfigure}{.24\textwidth}
    \centering
    \includegraphics[width=.95\linewidth]{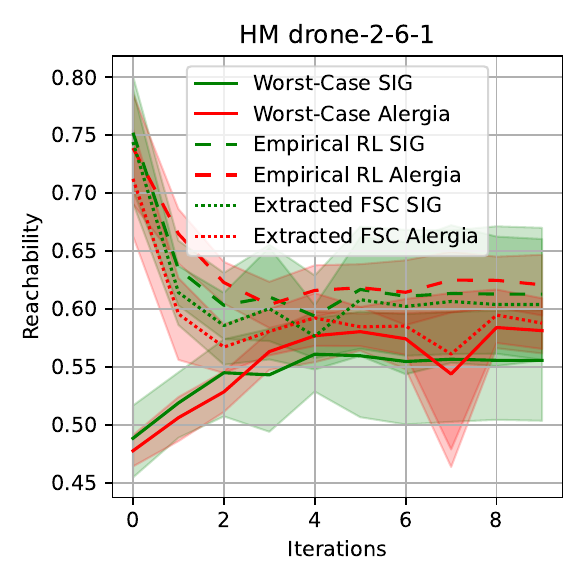}
    \caption{Drone-2-6-1}
    \end{subfigure}
    \begin{subfigure}{.24\textwidth}
    \centering
    \includegraphics[width=.95\linewidth]{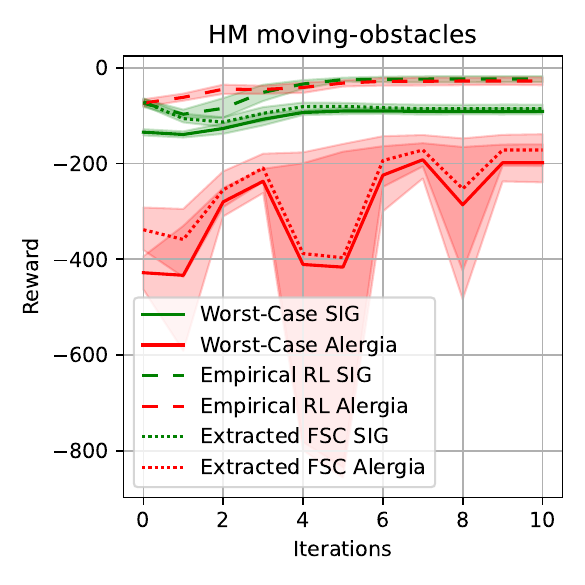}
    \caption{Moving-Obstacles}
    \end{subfigure}
    \begin{subfigure}{.24\textwidth}
    \centering
    \includegraphics[width=.95\linewidth]{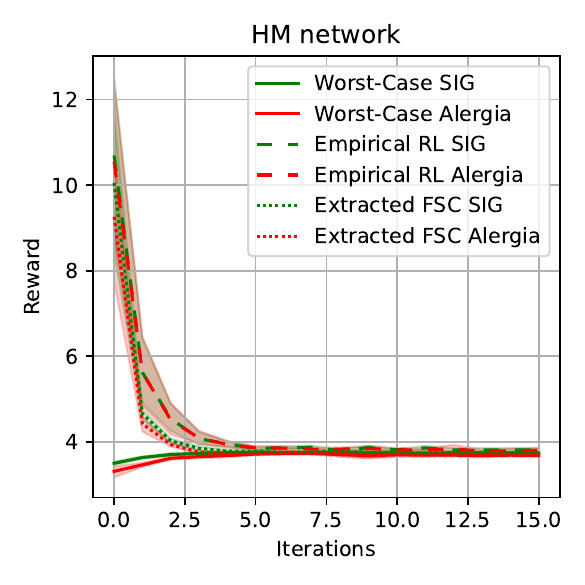}
    \caption{Network}
    \end{subfigure}
    \begin{subfigure}{.24\textwidth}
    \centering
    \includegraphics[width=.95\linewidth]{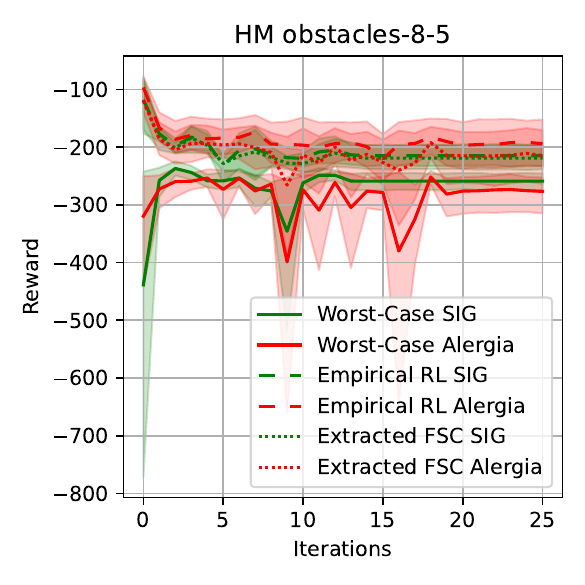}
    \caption{Obstacles-8-5}
    \end{subfigure}
    \begin{subfigure}{.24\textwidth}
    \centering
    \includegraphics[width=.95\linewidth]{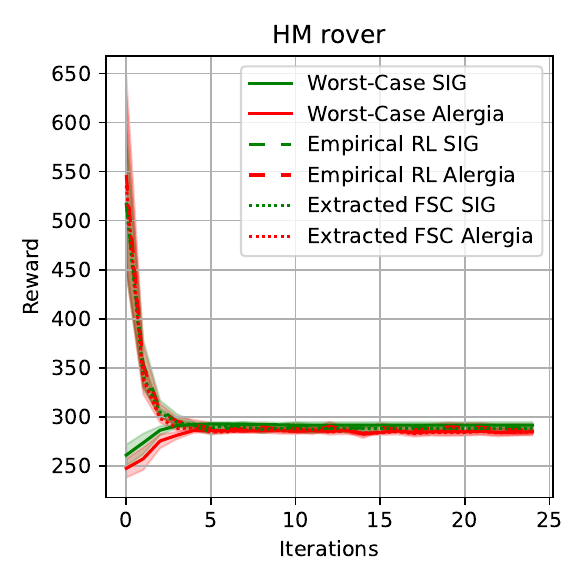}
    \caption{Rover}
    \end{subfigure}

    \caption{Figures for hidden model POMDPs. The green lines show the performance of our SIG method, while the red line shows the performance of Alergia. The solid lines show the verified performance, the dashed lines the empirical performance of the RL, and the dotted lines the empirical performance of the extracted controllers in a subset of POMDPs.}
    \label{fig:all_robust_pomdps}
    \Description{}
\end{figure}

\Cref{fig:all_robust_pomdps} shows our convergence curves for all hidden-model POMDP experiments. Those figures depict convergence curves of the empirical value over a subset of POMDPs over time, along with the performance of the extracted controllers in the same subset and their synthesized worst-case values.

\end{document}